\documentclass[11pt,a4paper]{article}
\usepackage[hyperref]{emnlp2020}
\usepackage{times}
\usepackage{latexsym}

\usepackage{microtype}
\usepackage{hyperref}
\usepackage{url}
\usepackage{subcaption} 
\usepackage{booktabs} %
\usepackage{enumitem}
\usepackage{graphicx}
\usepackage{bbm}
\usepackage{amsmath,amsfonts,bm}
\usepackage[T1]{fontenc}
\usepackage[utf8]{inputenc}
\usepackage{bbm}
\usepackage{amssymb}
\usepackage{algorithm}
\usepackage{algpseudocode}
\usepackage{algcompatible}
\usepackage{multirow}

\usepackage{mathtools}

\DeclareMathOperator*{\argmin}{arg\,min}
\aclfinalcopy %
\def\aclpaperid{2666} %

\setlist[enumerate,1]{leftmargin=1.2em,labelindent=0em,itemsep=0pt,labelsep*=0.5em}

\title{Cross-Lingual Text Classification with Minimal Resources\\by Transferring a Sparse Teacher}

\author{Giannis Karamanolakis, Daniel Hsu, Luis Gravano\\
Columbia University, New York, NY 10027, USA \\
\texttt{\{gkaraman, djhsu, gravano\}@cs.columbia.edu}
}

\date{}

\begin{document}
\maketitle
\begin{abstract}
Cross-lingual text classification alleviates the need for manually labeled documents in a target language by leveraging labeled documents from other languages. 
Existing approaches for transferring supervision across languages require \emph{expensive} cross-lingual resources, such as parallel corpora, while less expensive cross-lingual representation learning approaches train classifiers \emph{without} target labeled documents. 
In this work, we propose a cross-lingual teacher-student method, CLTS, that generates ``weak'' supervision in the target language using \emph{minimal} cross-lingual resources, in the form of a small number of word translations. 
Given a limited translation budget, CLTS extracts and transfers only the most important task-specific seed words across languages and initializes a teacher classifier based on the translated seed words.
Then, CLTS iteratively trains a more powerful student that also exploits the context of the seed words in \emph{unlabeled} target documents and outperforms the teacher. 
CLTS is simple and surprisingly effective in 18 diverse languages: by transferring just 20 seed words, even a bag-of-words logistic regression student outperforms state-of-the-art cross-lingual methods (e.g., based on multilingual BERT).
Moreover, CLTS can accommodate any type of student classifier: leveraging a \emph{monolingual} BERT student leads to further improvements and outperforms even more expensive approaches by up to 12\% in accuracy.
Finally, CLTS addresses emerging tasks in low-resource languages using just a small number of word translations.
\end{abstract}

\section{Introduction}
The main bottleneck in using supervised learning for multilingual document classification is the high cost of obtaining labeled documents for all of the target languages.
To address this issue in a target language $L_T$, we consider a cross-lingual text classification approach that requires labeled documents only in a source language $L_S$ and not in $L_T$.

\begin{figure}[t]
\centering
\includegraphics[scale=0.35]{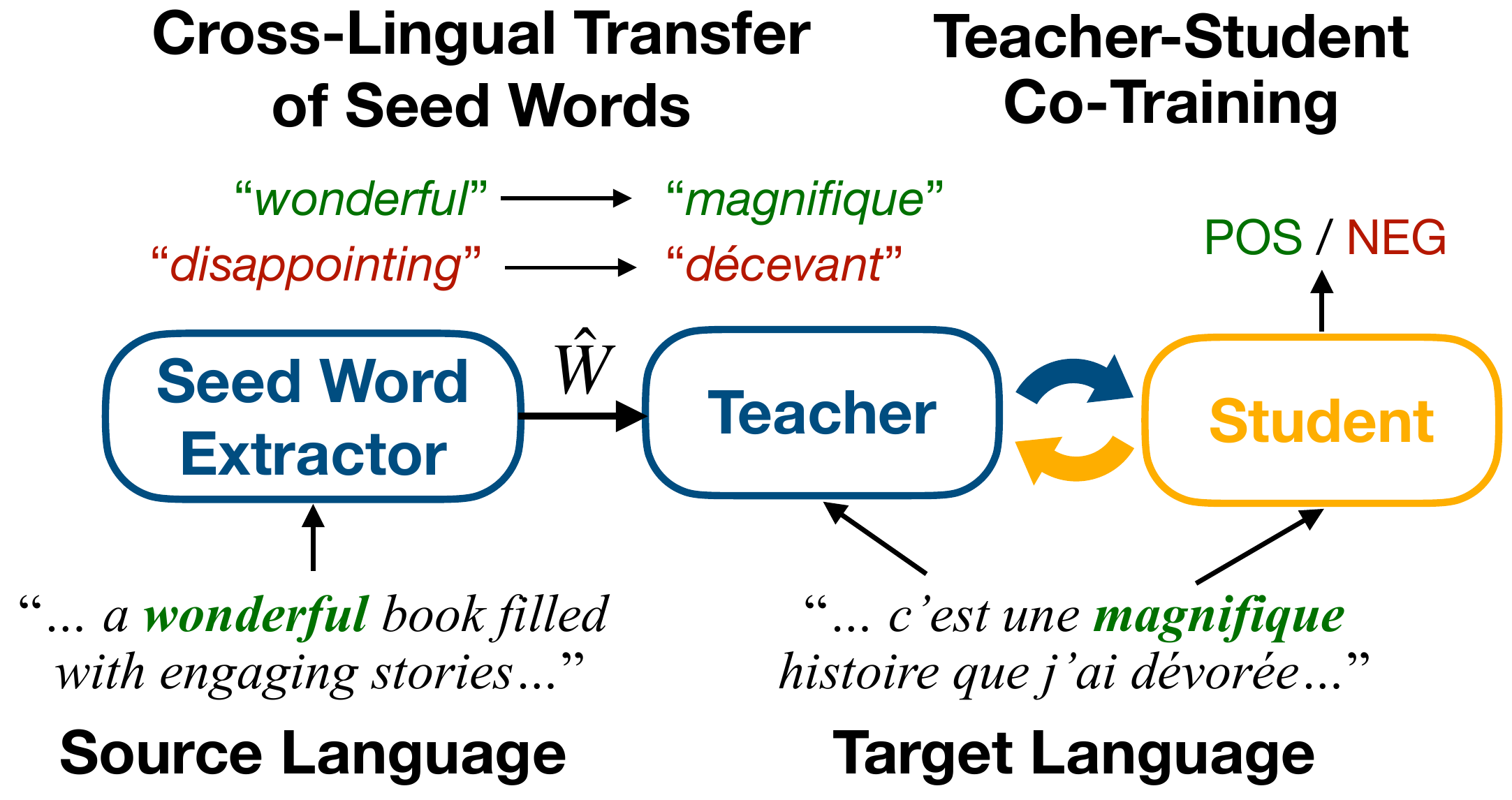}
\caption{Our cross-lingual teacher-student (CLTS) method trains a student classifier in the target language by transferring weak supervision across languages.}
\label{fig:CLTS-overview}
\end{figure}

Existing approaches for transferring supervision across languages rely on large parallel corpora or machine translation systems, which are expensive to obtain and are not available for many languages.\footnote{Google Translate (\url{https://translate.google.com/}) is available for 103 out of the about 4,000 written languages (\url{https://www.ethnologue.com/}).} 
 To scale beyond high-resource languages, multilingual systems have to reduce the cross-lingual requirements and operate under a limited budget of cross-lingual resources.
Such systems typically ignore target-language supervision, and rely on feature representations that bridge languages, such as cross-lingual word embeddings~\cite{ruder2019survey} or multilingual transformer models~\cite{wu2019beto, pires2019multilingual}.
This general approach is less expensive but has a key limitation: by not considering labeled documents in $L_T$, it may fail to capture predictive patterns that are specific to $L_T$.
Its performance is thus sensitive to the quality of pre-aligned features~\cite{glavavs2019properly}.

In this work, we show how to obtain weak supervision for training accurate classifiers in $L_T$ without using manually labeled documents in $L_T$ or expensive document translations.
We propose a novel approach for cross-lingual text classification that transfers weak supervision from $L_S$ to $L_T$ using \emph{minimal} cross-lingual resources: we only require a small number of task-specific keywords, or seed words, to be translated from $L_S$ to $L_T$.  
Our core idea is that the most indicative seed words in $L_S$ often translate to words that are also indicative in $L_T$.
For instance, the word ``wonderful” in English indicates positive sentiment, and so does its translation ``magnifique” in French. 
Thus, given a limited budget for word translations (e.g., from a bilingual speaker), only the most important seed words should be prioritized to transfer task-specific information from $L_S$ to $L_T$.

Having access only to limited cross-lingual resources creates important challenges, which we address with a novel cross-lingual teacher-student method, CLTS (see Figure~\ref{fig:CLTS-overview}). 
\paragraph{Efficient transfer of supervision across languages:}
As a first contribution, we present a method for cross-lingual transfer in low-resource settings with a limited word translation budget. 
CLTS extracts the most important seed words using the translation budget as a sparsity-inducing regularizer when training a classifier in $L_S$.
Then, it transfers seed words and the classifier's weights across languages, and initializes a teacher classifier in $L_T$ that uses the translated seed words.
\paragraph{Effective training of classifiers without using any labeled target documents:}
The teacher, as described above, predicts meaningful probabilities only for documents that contain translated seed words.
Because translations can induce errors and the translation budget is limited, the translated seed words may be noisy and not comprehensive for the task at hand.
As a second contribution, we extend the ``weakly-supervised co-training'' method of~\citet{karamanolakis2019cotraining} to our cross-lingual setting for training a stronger student classifier using the teacher and unlabeled-only target documents.
The student outperforms the teacher across all languages by 59.6\%.

\begin{figure}[t]
\centering
\includegraphics[width=\columnwidth]{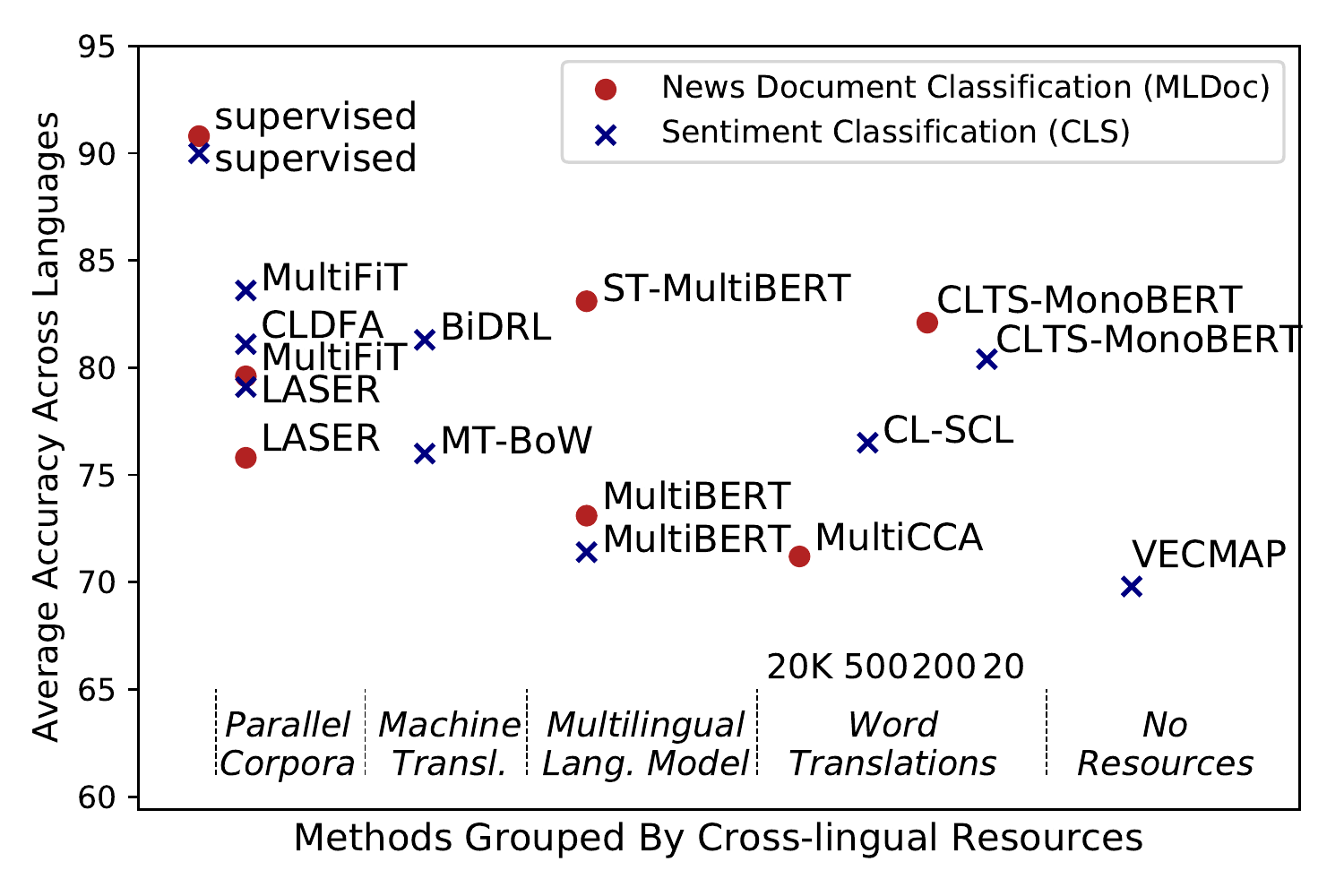}
\caption{%
CLTS leverages a small number of word translations more effectively than previous methods and sometimes outperforms more expensive methods. (Refer to Sections~\ref{s:related-work} and~\ref{s:experiments} for details.)}
\label{fig:CLTS-scatterplot}
\end{figure}

\paragraph{Robust performance across languages and tasks:}
As a third contribution, we empirically show the benefits of generating weak supervision in 18 diverse languages and 4 document classification tasks. 
With as few as 20 seed-word translations and a bag-of-words logistic regression student, CLTS outperforms state-of-the-art methods relying on more complex multilingual models, such as multilingual BERT, across most languages.
Using a monolingual BERT student leads to further improvements and outperforms even more expensive approaches (Figure~\ref{fig:CLTS-scatterplot}).
CLTS does not require cross-lingual resources such as parallel corpora, machine translation systems, or pre-trained multilingual language models, which makes it applicable in low-resource settings.
As a preliminary exploration, we address medical emergency situation detection in Uyghur and Sinhalese with just 50 translated seed words per language, which could be easily obtained from bilingual speakers. 

The rest of this paper is organized as follows. 
Section~\ref{s:related-work} reviews related work.
Section~\ref{s:problem-definition} defines our problem of focus.
Section~\ref{s:clts} presents our cross-lingual teacher-student approach.\footnote{Our Python implementation is publicly available at \href{https://github.com/gkaramanolakis/clts}{https://github.com/gkaramanolakis/clts}.} 
Section~\ref{s:experiments} describes our experimental setup and results. 
Finally, Section~\ref{s:conclusions} concludes and suggests future work. 

\section{Related Work}
\label{s:related-work}
Relevant work spans cross-lingual text classification and knowledge distillation.

\subsection{Cross-Lingual Text Classification}
\label{s:related-cross-lingual-text-classification}
We focus on a cross-lingual text classification scenario with labeled data in the source language $L_S$ and unlabeled data in the target language $L_T$.
We review the different types of required cross-lingual resources, starting with the most expensive types. 

\paragraph{Annotation Projection and Machine Translation.}
\label{ss:annotation-projection}
With parallel corpora (i.e., corpora where each document is written in both $L_S$ and $L_T$), 
a classifier trained in $L_S$ predicts labels for documents in $L_S$ and its predictions are projected to documents in $L_T$ to train a classifier in $L_T$~\cite{mihalcea2007learning,rasooli2018cross}.
Unfortunately, parallel corpora are hard to find, especially in low-resource domains and languages.

Without parallel corpora, documents can be translated using machine translation (MT) systems~\cite{wan2008using,wan2009co,salameh2015sentiment,mohammad2016translation}. 
However, high-quality MT systems are limited to high-resource languages.
Even when an MT system is available, translations may change document semantics and degrade classification accuracy~\cite{salameh2015sentiment}.

\paragraph{Cross-Lingual Representation Learning.}
Other approaches rely on less expensive resources to align feature representations across languages, typically in a shared feature space to enable cross-lingual model transfer.
Cross-lingual word embeddings, or CLWE, represent words from different languages in a joint embedding space, where words with similar meanings obtain similar vectors regardless of their language. 
(See~\citet{ruder2019survey} for a survey.)
Early CLWE approaches required expensive parallel data~\cite{klementiev2012inducing,tackstrom2012cross}.  %
In contrast, later approaches rely on high-coverage bilingual dictionaries~\cite{gliozzo2006exploiting,faruqui2014improving,gouws2015bilbowa} or smaller ``seed'' dictionaries~\cite{gouws2015simple, artetxe2017learning}. 
Some recent CLWE approaches require no cross-lingual resources~\cite{conneau2017word, artetxe2018robust, chen2018unsupervised, sogaard2018limitations} but perform substantially worse than approaches using seed dictionaries of 500-1,000 pairs~\cite{vulic2019we}.
Our approach does not require CLWE and achieves competitive classification performance with substantially fewer translations of \emph{task-specific} words. %

Recently, multilingual transformer models were pre-trained in multiple languages in parallel using language modeling objectives~\cite{devlin2019bert,conneau2019cross}.
Multilingual BERT, a version of BERT~\cite{devlin2019bert} that was trained on 104 languages in parallel without using any cross-lingual resources, has received significant attention~\cite{wang2019cross, singh2019bert,rogers2020primer}. 
Multilingual BERT performs well on zero-shot cross-lingual transfer~\cite{wu2019beto,pires2019multilingual} and its performance can be further improved by considering target-language documents through self-training~\cite{dong2019robust}.
In contrast, our approach does not require multilingual language models and sometimes outperforms multilingual BERT using a \emph{monolingual} BERT student. %

\subsection{Knowledge Distillation}
\label{s:related-knowledge-distillation}
Our teacher-student approach is related to ``knowledge distillation''~\cite{bucilua2006model,ba2014deep,hinton2015distilling}, where a student classifier is trained using the predictions of a teacher classifier.
\citet{xu2017cross} apply knowledge distillation for cross-lingual text classification but require expensive parallel corpora. 
MultiFiT~\cite{eisenschlos2019multifit} trains a classifier in $L_T$ using the predictions of a cross-lingual model, namely, LASER~\cite{artetxe2019massively}, that also requires large parallel corpora.
\citet{vyas2019weakly} classify the semantic relation (e.g., synonymy) between two words from different languages by transferring \emph{all} training examples across languages. 
Our approach addresses a different problem, where training examples are full documents (not words), and transferring source training documents would require MT. 
Related to distillation is the semi-supervised approach of~\citet{shi2010cross} that trains a target classifier by transferring a source classifier using high-coverage dictionaries.
Our approach is similar, but trains a classifier using sparsity regularization, and translates only the most important seed words.

\section{Problem Definition}
\label{s:problem-definition}
Consider a source language $L_S$, a target language $L_T$, and a classification task with $K$ predefined classes of interest $\mathcal{Y} = \{1, \dots, K\}$ (e.g., sentiment categories).
Labeled documents $D_S = \{(x^{S}_i, y_i)\}_{i=1}^N$ are available in $L_S$, where $y_i \in \mathcal{Y}$ and each source document $x^S_i$ is a sequence of words from the source vocabulary $V_S$.
Only unlabeled documents $D_T = \{x^T_i\}_{i=1}^M$ are available in $L_T$, where each target document $x^T_i$ is a sequence of words from the target vocabulary $V_T$.
We assume that there is no significant shift in the conditional distribution of labels given documents across languages.
Furthermore, we assume a limited translation budget, so that up to $B$ words can be translated from $L_S$ to $L_T$.

Our goal is to use the labeled source documents $D_S$, the unlabeled target documents $D_T$, and the translations of no more than $B$ source words to train a classifier that, given an unseen test document $x^T_i$ in the target language $L_T$, predicts the corresponding label $y_i \in \mathcal{Y}$.

\section{Cross-Lingual Teacher-Student}
\label{s:clts}
We now describe our cross-lingual teacher-student method, CLTS, for cross-lingual text classification. 
Given a limited budget of $B$ translations, CLTS extracts only the $B$ most important seed words in $L_S$ (Section~\ref{s:source-extraction}).
Then, CLTS transfers the seed words and their weights from $L_S$ to $L_T$, to initialize a classifier in $L_T$ (Section~\ref{s:cross-lingual-transfer}).
Using this classifier as a teacher, CLTS trains a student that predicts labels using both seed words and their context in target documents (Section~\ref{s:target-bootstrapping}).  

\subsection{Seed-Word Extraction in $L_S$}
\label{s:source-extraction}
CLTS starts by automatically extracting a set $G_k^S$ of indicative seed words per class $k$ in $L_S$.
Previous extraction approaches, such as tf-idf variants~\cite{angelidis2018summarizing}, have been effective in monolingual settings with limited labeled data.
In our scenario, with sufficiently many labeled \emph{source} documents and a limited translation budget $B$, we propose a different approach based on a supervised classifier trained with sparsity regularization.

Specifically, CLTS extracts seed words from the weights $W \in \mathbb{R}^{K \times |V_S|}$ of a classifier trained using $D_S$. 
Given a source document $x_i^S$ with a bag-of-words encoding $h_i^S \in \mathbb{R}^{|V_S|}$, the classifier predicts class probabilities $p_i=\langle p_i^1, \dots, p_i^K \rangle = \operatorname{softmax}(Wh_i)$.
CLTS includes the word $v_c \in V_S$ in $G_k^S$ if the classifier considers it to increase the probability $p_i^k$ through a positive weight $W_{kc}$:
\begin{equation}
    \label{eq:seed-word-extraction}
    G_k^S = \{v_c^S \mid W_{kc}>0\}.
\end{equation}
The set of all source seed words $G^S = G_1^S \cup \dots \cup G_K^S$ may be much larger than the translation budget $B$. 
We encourage the classifier to capture only the most important seed words \emph{during} training through sparsity regularization:
\begin{equation}
\label{eq:supervised-training}
   \hat W =  \argmin_{W} \sum_{i=1}^N \mathcal{L}(y_{i}, W h_i^{S}) + \lambda_B \mathcal{R}_{\text{sparse}}(W)
\end{equation}
where $\mathcal{L}$ is the training loss function (logistic loss), $\mathcal{R}_{\text{sparse}}(.)$ is a sparsity regularizer (L1 norm), and $\lambda_B \in \mathbb{R}$ is a hyperparameter controlling the relative power of $\mathcal{R}_{\text{sparse}}$. Higher $\lambda_B$ values lead to sparser matrices $\hat W$ and thus to fewer seed words. 
Therefore, we tune\footnote{We efficiently tune $\lambda_B$ by computing the ``regularization path'' with the ``warm-start'' technique~\cite{koh2007interior}.} $\lambda_B$ to be as high as possible while at the same time leading to the extraction of at least $B$ seed words. 
After training, $G^S$ consists of the $B$ seed words with highest weight.

\begin{figure}[t]
\centering
\includegraphics[scale=0.3]{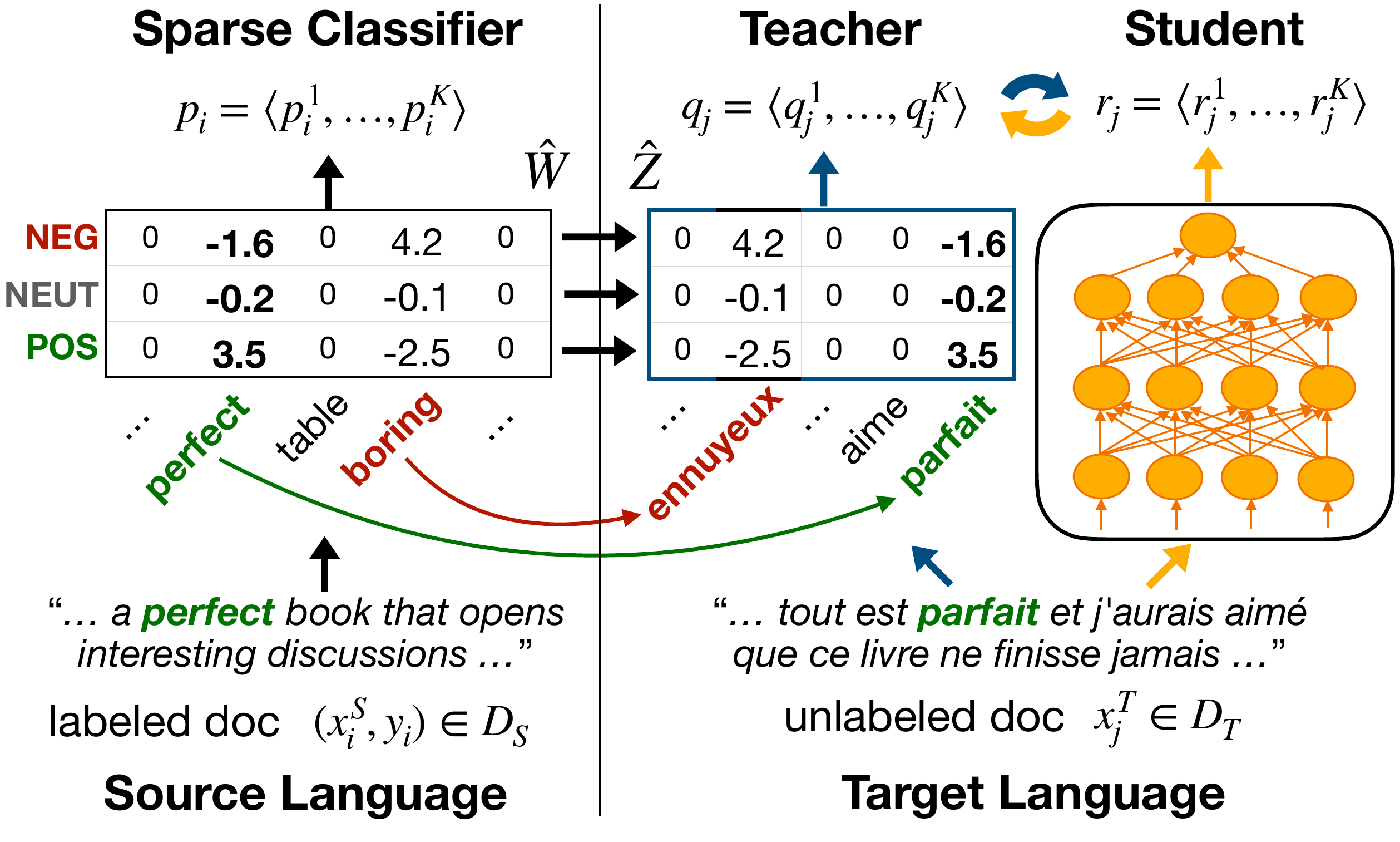}
	\caption{CLTS (1) learns a sparse weight matrix $\hat W$ in $L_S$; (2) transfers the columns of $\hat W$ for $B$ seed words to initialize $\hat Z$; and (3) uses $\hat Z$ as a teacher to iteratively train a student on unlabeled documents $D_T$.}
\label{fig:clts_architecture}
\end{figure}

\subsection{Cross-Lingual Seed Weight Transfer}
\label{s:cross-lingual-transfer}
We now describe our cross-lingual transfer method.
CLTS transfers both translated seed words and their learned weights to initialize a ``weak'' classifier in $L_T$ that considers translated seed words and their relative importance for the target task. 

Specifically, CLTS first translates the $B$ seed words in $G^S$ into a set $G^T$ with seed words in $L_T$.
Then, for each translation pair $(v^S, v^T)$, CLTS transfers the column for $v^S$ in $\hat W$ to a corresponding column for $v^T$ in a $K \times |V_T|$ matrix $\hat Z$:
\begin{equation}
    \label{eq:weight-transfer}
    \hat Z_{k,v^S} = \hat W_{k, v^T} \enskip \forall k\in[K]
\end{equation}
Importantly, for each word, we transfer the weights for all classes (instead of just a single weight $\hat W_{kc}$) across languages.
Therefore, \emph{without using any labeled documents} in $L_T$, CLTS constructs a classifier that, given a test document $x_j^T$ in $L_T$, predicts class probabilities $q_j=\langle q_j^1, \dots, q_j^K \rangle$: 
\begin{equation}
    \label{eq:teacher-pred}
    q_j^k = \frac{\exp{(\hat z_k^\top h_j^T})}{\sum_{k'} \exp{(\hat z_{k'}^\top h_j^T)}},
\end{equation}
where $h_j^T \in \mathbb{R}^{|V_T|}$ is a bag-of-words encoding for $x_j^T$ and $\hat z_k$ is the $k$-th row of $\hat Z$.
Note that columns of $\hat Z$ for non-seed words in $V_T$ are all zeros and thus this classifier predicts meaningful probabilities only for documents with seed words in $G^T$.

\subsection{Teacher-Student Co-Training in $L_T$}
\label{s:target-bootstrapping}
We now describe how CLTS trains a classifier in $L_T$ that leverages indicative features, which may not be captured by the small set of translated seed words.
As illustrated in Figure~\ref{fig:clts_architecture}, translated seed words (e.g., ``parfait'') often co-occur with other words (e.g., ``aime,'' meaning ``love'') that have zero weight in $\hat Z$ but are also helpful for the task at hand.
To exploit such words in the absence of labeled target documents, we extend the monolingual weakly-supervised co-training method of~\citet{karamanolakis2019cotraining} to our cross-lingual setting, and use our classifier based on translated seed words as a teacher to train a student, as we describe next. 

First, CLTS uses our classifier from Equation~\ref{eq:student-pred} as a teacher to predict labels $q_j$ for \emph{unlabeled} documents $x_j^T \in D_T$ that contain seed words: $D_T' =\{(x_j^T,q_j)\}_{x_j^T \mid x_j^T \cap G^T \neq \varnothing} \subseteq D_T$.
Note that our teacher with weights transferred across languages is different than that of~\citet{karamanolakis2019cotraining}, which simply ``counts'' seed words. 

Next, CLTS trains a student $f^T$ that also exploits the context of the seed words.
Given a document $x_j^T$ in $L_T$, the student predicts class probabilities: 
\begin{equation}
r_j = \langle r_j^1, \dots, r_j^K \rangle = f^T(x_j^T; \theta),
\label{eq:student-pred}
\end{equation}
where the predictor function $f^T$ with weight parameters $\theta$ can be of any type, such as a pre-trained transformer-based classifier that captures language-specific word composition.
The student is trained via the ``distillation'' objective: %
\begin{equation}
\label{eq:distillation-loss}
   \hat \theta =  \argmin_{\theta} \sum_{(x_j^T, q_j) \in D_T'} H(q_j, f^T(x_j^T)) + \lambda \mathcal{R}(\theta),
\end{equation}
where $H(q, r) = - \sum_{k} q^k \log r^k$ is the cross entropy between student's and teacher's predictions, $\mathcal{R}(.)$ is a regularizer (L2 norm),  and $\lambda \in \mathbb{R}$ is a hyperparameter controlling the relative power of $\mathcal{R}$.
Importantly, through extra regularization ($\mathcal{R}$, dropout) the student also associates non-seed words with target classes, and generalizes better than the teacher by making predictions even for documents that do not contain any seed words.

Then, CLTS uses the student in place of the teacher to annotate \emph{all} $M$ unlabeled examples in $D_T$ and create $D_T'=\{(x_j^T, \hat f^T(x_j^T)\}_{j\in[M]}$.
While in the first iteration $D_T'$ contains only documents with seed words, in the second iteration CLTS adds in $D_T'$ \emph{all} unlabeled documents to create a larger training set for the student. 
This also differs from~\citet{karamanolakis2019cotraining}, which updates the weights of the initial seed words but does not provide pseudo-labels for documents with no seed words. 
This change is important in our cross-lingual setting with a limited translation budget, where the translated seed words $G^T$ may only cover a very small subset $D_T'$ of $D_T$. 

\begin{figure}
    \centering
    \begin{algorithm}[H]
\caption{Cross-Lingual Teacher-Student}
\begin{algorithmic}[1]
\Statex{\textbf{Input:} Unlabeled documents $D_T = \{x^T_j\}_{j=1}^M$, labeled documents $D_S = \{(x^{S}_i, y_i)\}_{i=1}^N$, budget of up to $B$ word translations ($L_S$ to $L_T$)} 
\Statex{\textbf{Output:} $\hat f^T$: predictor function in $L_T$}
    \State{Learn $\lambda_B$-sparse $\hat W$ using $D_S$, $B$ (Eq.~\eqref{eq:supervised-training})}
    \State{Extract $B$ seed words $G^S$ from $\hat W$ (Eq.~\eqref{eq:seed-word-extraction})}
    \State{Translate $G^S$ to target seed words $G^T$ in $L_T$}
    \State{Transfer $\hat W$ to initialize teacher $\hat Z$ (Eq.~\eqref{eq:weight-transfer})}
    \State{Get $D_T' =\{(x_j^T,q_j)\}_{x_j^T \mid x_j^T \cap G^T \neq \varnothing}$ (Eq.~\eqref{eq:teacher-pred})}
    \State{\textbf{Repeat until convergence}}
      \begin{algorithmic}
      \State{a. Learn student $\hat f^T$ using $D_T'$ (Eq.~\eqref{eq:distillation-loss})}
      \State{b. Get $D_T'=\{(x_j^T, \hat f^T(x_j^T)\}_{j\in[M]}$ (Eq.~\eqref{eq:student-pred})}
    \end{algorithmic}
\end{algorithmic}
\end{algorithm}
\end{figure}

Algorithm 1 summarizes the CLTS method for cross-lingual classification by translating $B$ seed words. 
Iterative co-training converges when the disagreement between the student’s and teacher’s hard predictions on unlabeled data stops decreasing.
In our experiments, just two rounds of co-training are generally sufficient for the student to outperform the teacher and achieve competitive performance even with a tight translation budget $B$. 

\section{Experiments}
\label{s:experiments}
We now evaluate CLTS for several cross-lingual text classification tasks in various languages.  

\subsection{Experimental Settings}
\paragraph{Datasets:}
We use English (En) as a source language, and evaluate CLTS on 18 diverse target languages: Bulgarian (Bg), German (De), Spanish (Es), Persian (Fa), French (Fr),  Croatian (Hr), Hungarian (Hu), Italian (It), Japanese (Ja), Polish (Pl), Portuguese (Pt), Russian (Ru), Sinhalese (Si), Slovak (Sk), Slovenian (Sl), Swedish (Sv), Uyghur (Ug), and Chinese (Zh).
We focus on four classification tasks: \textbf{T1:} 
4-class classification of news documents in the MLDoc corpus~\cite{schwenk2018corpus};
\textbf{T2:} binary sentiment classification of product reviews in the CLS corpus~\cite{prettenhofer2010cross}; \textbf{T3:} 3-class sentiment classification of tweets in the Twitter Sentiment corpus (TwitterSent;~\citet{mozetic2016}), Persian reviews in the SentiPers corpus~\cite{hosseini2018sentipers}, and Uyghur documents in the LDC LORELEI corpus~\cite{strassel2016lorelei}; 
and \textbf{T4:} medical emergency situation detection in Uyghur and Sinhalese documents from the LDC LORELEI corpus.
The appendix discusses additional dataset details.

\paragraph{Experimental Procedure:}
We use English as the source language, where we train a source classifier and extract $B$ seed words using labeled documents (Section~\ref{s:source-extraction}). 
 Then, we obtain translations for $B\leq 500$ English seed words using the MUSE\footnote{\url{https://github.com/facebookresearch/MUSE\#ground-truth-bilingual-dictionaries}} bilingual dictionaries~\cite{conneau2017word}.
 For Uyghur and Sinhalese, which have no entries in MUSE, we translate seed words through Google Translate.\footnote{Google Translate started supporting Uyghur on February 26, 2020, and Sinhalese at an earlier (unknown) time.}
 The appendix reports additional seed-word translation details.
We do not use labeled documents in the target language for training (Section~\ref{s:problem-definition}).
We report both the teacher's and student's performance in $L_T$ averaged over 5 different runs. 
We consider any test document that contains no seed words as a ``mistake'' for the teacher.

\paragraph{Model Configuration:}
For the student, we experiment with a bag-of-ngrams ($n=1,2$) logistic regression classifier (LogReg) and a linear classifier using pre-trained monolingual BERT embeddings as features (MonoBERT;~\citet{devlin2019bert}).
The appendix lists the implementation details. 
We do not optimize any hyperparameters in the target language, except for $B$, which we vary between $6$ and $500$ to understand the impact of translation budget on performance.
CLS does not contain validation sets, so we fix $B=20$ and translate 10 words for each of the two sentiment classes.
More generally, to cover all classes we extract and translate $\frac{B}{K}$ seed words per class.
We perform two rounds of teacher-student co-training, which provided most of the improvement in~\citet{karamanolakis2019cotraining}. 

\paragraph{Model Comparison:}
For a robust evaluation of CLTS, we compare models with different types of cross-lingual resources. %
\textbf{\textit{Project-*}} uses the parallel LDC or EuroParl (EP) corpora for annotation projection~\cite{rasooli2018cross}.
\textbf{\textit{LASER}} uses millions of parallel corpora to obtain cross-lingual sentence embeddings~\cite{artetxe2019massively}.
\textbf{\textit{MultiFiT}} uses \textbf{\textit{LASER}} to create pseudo-labels in $L_T$~\cite{eisenschlos2019multifit} and trains a classifier in $L_T$ based on a pre-trained language model~\cite{howard2018universal}. 
\textbf{\textit{CLWE-par}} uses parallel corpora to train CLWE~\cite{rasooli2018cross}.
\textbf{\textit{MT-BOW}} uses Google Translate to translate test documents from $L_T$ to $L_S$ and applies a bag-of-words classifier in $L_S$~\cite{prettenhofer2010cross}.
\textbf{\textit{BiDRL}} uses Google Translate to translate documents from $L_S$ to $L_T$ and $L_T$ to $L_S$~\cite{zhou2016cross}.
\textbf{\textit{CLDFA}} uses task-specific parallel corpora for cross-lingual distillation~\cite{xu2017cross}.
\textbf{\textit{SentiWordNet}} uses bilingual dictionaries with over 20K entries to transfer the SentiWordNet03~\cite{baccianella2010sentiwordnet} to the target language and applies a rule-based heuristic~\cite{rasooli2018cross}.
\textbf{\textit{CLWE-Wikt}} uses bilingual dictionaries with over 20K entries extracted from Wiktionary\footnote{\url{https://www.wiktionary.org/}} to create CLWE for training a bi-directional LSTM classifier~\cite{rasooli2018cross}. 
\textbf{\textit{MultiCCA}} uses bilingual dictionaries with around 20K entries to train CLWE~\cite{ammar2016massively}, trains a convolutional neural network (CNN) in $L_S$ and applies it in $L_T$~\cite{schwenk2018corpus}.
\textbf{\textit{CL-SCL}} obtains 450 word translations as ``pivots'' for cross-lingual domain adaptation~\cite{prettenhofer2010cross}. 
Our CLTS approach uses $B$ word translations not for domain adaptation but to create weak supervision in $L_T$ through the teacher (Teacher) for training the student (Student-LogReg or Student-MonoBERT).
\textbf{\textit{VECMAP}} uses identical strings across languages as a weak signal to train CLWE~\cite{artetxe2017learning}. %
\textbf{\textit{MultiBERT}} uses multilingual BERT to train a classifier in $L_S$ and applies it in $L_T$~\cite{wu2019beto} without considering labeled documents in $L_T$ (zero-shot setting).
\textbf{\textit{ST-MultiBERT}} further considers labeled documents in $L_T$ for fine-tuning multilingual BERT through self-training~\cite{dong2019robust}.
The appendix discusses more comparisons.

\begin{figure*}[t]
    \centering
    \begin{subfigure}[t]{0.62\textwidth}
\resizebox{1\columnwidth}{!}{
\begin{tabular}{|l|ccccccc|c|}
\hline
 \textbf{Method} & \textbf{De}   & \textbf{Es}   & \textbf{Fr}   & \textbf{It}   & \textbf{Ru}   & \textbf{Zh}   & \textbf{Ja}   & \textbf{AVG}  \\\hline 
\multicolumn{9}{c}{\textit{Methods below use parallel corpora (MultiFiT requires LASER)}}\\\hline 
LASER & 87.7 & \textbf{79.3} & 84.0 & 71.2 & 67.3 & 76.7 & 64.6 & 75.8\\
MultiFiT & \textbf{91.6} &	79.1 &	\textbf{89.4}&	\textbf{76.0} &	\textbf{67.8} & \textbf{82.5} &	\textbf{69.6} &	\textbf{79.4} \\\hline\hline
\multicolumn{9}{c}{\textit{Methods below use pre-trained multi-lingual language models}}\\\hline
MultiBERT & 79.8 & 72.1 & 73.5 & 63.7 & 73.7 & 76.0 & 72.8 & 73.1 \\
ST-MultiBERT & \textbf{90.0} & \textbf{85.3} & \textbf{88.4} & \textbf{75.2}  & \textbf{79.3} & \textbf{87.0} & \textbf{76.8}  & \textbf{83.1} \\\hline 
\multicolumn{9}{c}{\textit{Methods below use bilingual dictionaries (Student requires Teacher)}}\\\hline 
MultiCCA ($B$=20K) &81.2 & 72.5 & 72.4 & 69.4 & 60.8 & 74.7 & 67.6 & 71.2 \\
Teacher ($B$=160) &  72.7 & 73.5 & 77.6 & 62.5 & 46.9 & 53.3 & 31.9 & 59.8 \\
Student-LogReg & 87.4 & 86.0 & 89.1 & 70.5 & 71.9 & 82.4 & 68.8 & 79.4 \\
Student-MonoBERT  & \textbf{90.4} & \textcolor{red}{\textbf{86.3}} & \textcolor{red}{\textbf{91.2}} & \textbf{74.7} & \textbf{75.6} & \textbf{84.0} & \textbf{72.6} & \textbf{82.1}\\

\hline 
\end{tabular}}
\caption{Accuracy results on MLDoc.}
\label{tab:mldoc-results-full}
    \end{subfigure}%
    ~ 
    \begin{subfigure}[t]{0.38\textwidth}
        \centering
\resizebox{1\columnwidth}{!}{
\begin{tabular}{|l|lll|l|}
\hline 
 \textbf{Model}             & \textbf{De}   & \textbf{Fr}   & \textbf{Ja}   & \textbf{AVG}  \\\hline
\multicolumn{5}{c}{\textit{Methods below use parallel corpora or MT}}\\\hline   
MT-BOW        & 78.3 & 78.5 & 71.2 & 76.0 \\
BiDRL         & 84.3 & 83.5 & 76.2 & 81.3 \\
CLDFA         & 82.0 & 83.1 & 78.1 & 81.1 \\
LASER         & 80.4 & 82.7 & 75.3 & 79.5 \\
MultiFiT      & \textbf{85.3} & \textbf{85.6} & \textbf{79.9} & \textbf{83.6} \\\hline
\multicolumn{5}{c}{\textit{Methods below use multi-lingual language models}}\\\hline
MultiBERT     & 72.0 & 75.4 & 66.9 & 71.4 \\\hline 
\multicolumn{5}{c}{\textit{Methods below use dictionaries or no resources}}\\\hline
VECMAP        & 75.3 & 78.2 & 55.9 & 69.8 \\
CL-SCL ($B$=450) & 78.1 & 78.4 & 73.1 & 76.5 \\
Teacher ($B$=20) & 38.1 & 48.6 & 22.7 & 36.5 \\
Student-LogReg     & 78.7 & 79.6 & \textbf{78.6} & 79.0 \\
Student-MonoBERT   & \textbf{80.1} & \textbf{83.4} & 77.6 & \textbf{80.4}\\\hline 
\end{tabular}}
\caption{Accuracy results on CLS.}
\label{tab:cls-results-full}
\end{subfigure}

\begin{subfigure}[t]{1\textwidth}
    \centering 
    \resizebox{1\columnwidth}{!}{
    \begin{tabular}{|l|cccccccccccccc|c|}
\hline 
Method & \textbf{Ar}   & \textbf{Bg}   & \textbf{De}   & \textbf{Es}   & \textbf{Fa}   & \textbf{Hr}   & \textbf{Hu}   & \textbf{Pl}   & \textbf{Pt}   & \textbf{Ru}   & \textbf{Sk}   & \textbf{Sl}   & \textbf{Sv}  & \textbf{Ug} & \textbf{AVG}  \\\hline
\multicolumn{16}{c}{\textit{Methods below use parallel corpora}}\\\hline 
Project-LDC & 37.2 & -&-& \textbf{42.7} & 33.1 &-& \textbf{47.0}   &- &-& \textbf{48.0}   &-&  -   &- & \textbf{38.6} &(41.1)\\
Project-EP & - &\textbf{38.7} &	\textbf{47.3} &	41.8 &	- &	-& 38.1 &	38.8 &	\textbf{39.3} &-&\textbf{30.0}&	\textbf{44.6}&	\textbf{44.6}&-&(40.4)\\
CLWE-Par   & \textbf{37.3} & 33.0   & 43.5 & 42.6 & \textbf{40.1} & \textbf{30.8} & 41.1 & \textbf{41.7} & 38.6 & 44.8 & 22.6 & 32.2 & 39.1 & 30.0 &37.0 \\\hline 
\multicolumn{16}{c}{\textit{Methods below use comparable corpora or bilingual dictionaries}}\\\hline 
CLWE-CP  & 21.1 & 28.6 & 37.7 & 27.7 & 20.7 & 13.9 & 22.4 & 30.2 & 22.2 & 25.3 & 24.6 & 25.3 & 31.1 & 25.7 & 25.5 \\
SentiWordNet ($B$>20K) & 25.6 & 30.6 & 32.0   & 25.3 & 25.3 & 19.8 & 29.2 & 26.0   & 22.9 & 29.5 & 19.2 & 28.1 & 22.7 & 36.7 & 26.6 \\
CLWE-Wikt ($B$>20K) & 31.0   & 45.3 & 51.0   & 37.7 & 31.7 & -    & 40.8 & 32.9 & 35.4 & \textbf{43.8} & 36.6 & 32.1 & 40.4 &   28.0 & (37.4) \\
Teacher ($B$=500) & 22.7 & 42.8 & 45.5 & 42.7 & 30.9 & 36.4 & 39.4 & 40.7 & 34.4 & 29.8 & 40.4 & 29.5 & 38.7 & 20.3 &35.3 \\
Student-LogReg & \textcolor{red}{\textbf{39.0}}   & \textcolor{red}{\textbf{46.3}} & \textcolor{red}{\textbf{52.5}} & \textcolor{red}{\textbf{44.9}} & \textcolor{red}{\textbf{45.7}} & \textcolor{red}{\textbf{39.4}} & \textbf{45.2} & \textcolor{red}{\textbf{45.4}} & \textbf{38.7} & 43.2 & \textcolor{red}{\textbf{43.3}} & \textbf{42.1} & \textcolor{red}{\textbf{50.4}} & \textcolor{red}{\textbf{41.2}} &\textcolor{red}{\textbf{44.1}}\\
\hline
\end{tabular}}
\caption{Macro-averaged F1 results on TwitterSent, SentiPers, and LORELEI.}
\label{tab:twittersent-results-full}
\end{subfigure}
\caption{Classification results, with methods grouped according to the type of cross-lingual resources required. For some methods, average performance (rightmost column) is in parentheses because it is computed on a subset of languages.
Across all datasets, CLTS outperforms other methods that require similar types of cross-lingual resources; in many cases (red) CLTS outperforms even \emph{more expensive} state-of-the-art approaches.}
\label{fig:all-results}
\end{figure*}
\subsection{Experimental Results}
\label{s:experimental-results}
Figure~\ref{fig:all-results} shows results for each classification task and language. 
The rightmost column of each table reports the average performance across all languages (and domains for CLS).
For brevity, we report the average performance across the three review domains (Books, DVD, Music) for each language in the CLS corpus. 
The appendix discusses detailed results and ablation experiments.

\paragraph{Student outperforms Teacher.}
Teacher considers the noisy translated seed words for classification. 
Even the simple Student-LogReg technique leverages the context of the seed words and substantially outperforms Teacher. 
Leveraging pre-trained representations in Student-MonoBERT leads to further improvement.
On average, across all languages and datasets, Student outperforms Teacher by 59.6\%: CLTS effectively improves performance in $L_T$ \emph{without using labeled documents}.

\paragraph{Student outperforms previous approaches.}
Student-MonoBERT outperforms \textit{MultiBERT} by 12.5\% on average across all languages and domains in MLDoc and CLS: CLTS effectively generates weak supervision in $L_T$ for fine-tuning monolingual BERT.
Importantly, CLTS is effective under minimal resources: with the translation of just $\frac{B}{K}$ seed words per class, Student-LogReg outperforms other approaches that rely on much larger dictionaries (\textit{MultiCCA}, \textit{CL-SCL}, \textit{SentiWordNet}, \textit{CLWE-Wiktionary}).
Surprisingly, in several languages CLTS outperforms even more expensive approaches that rely on parallel corpora or machine translation systems (\textit{LASER}, \textit{MultiFiT}, \textit{MT-BOW}, \textit{BiDRL}, \textit{CLDFA}, \textit{CLWE-BW}, \textit{Project-LDC}).

\begin{figure}[t]
\centering
\includegraphics[height = 4.8cm, width = 7.8cm]{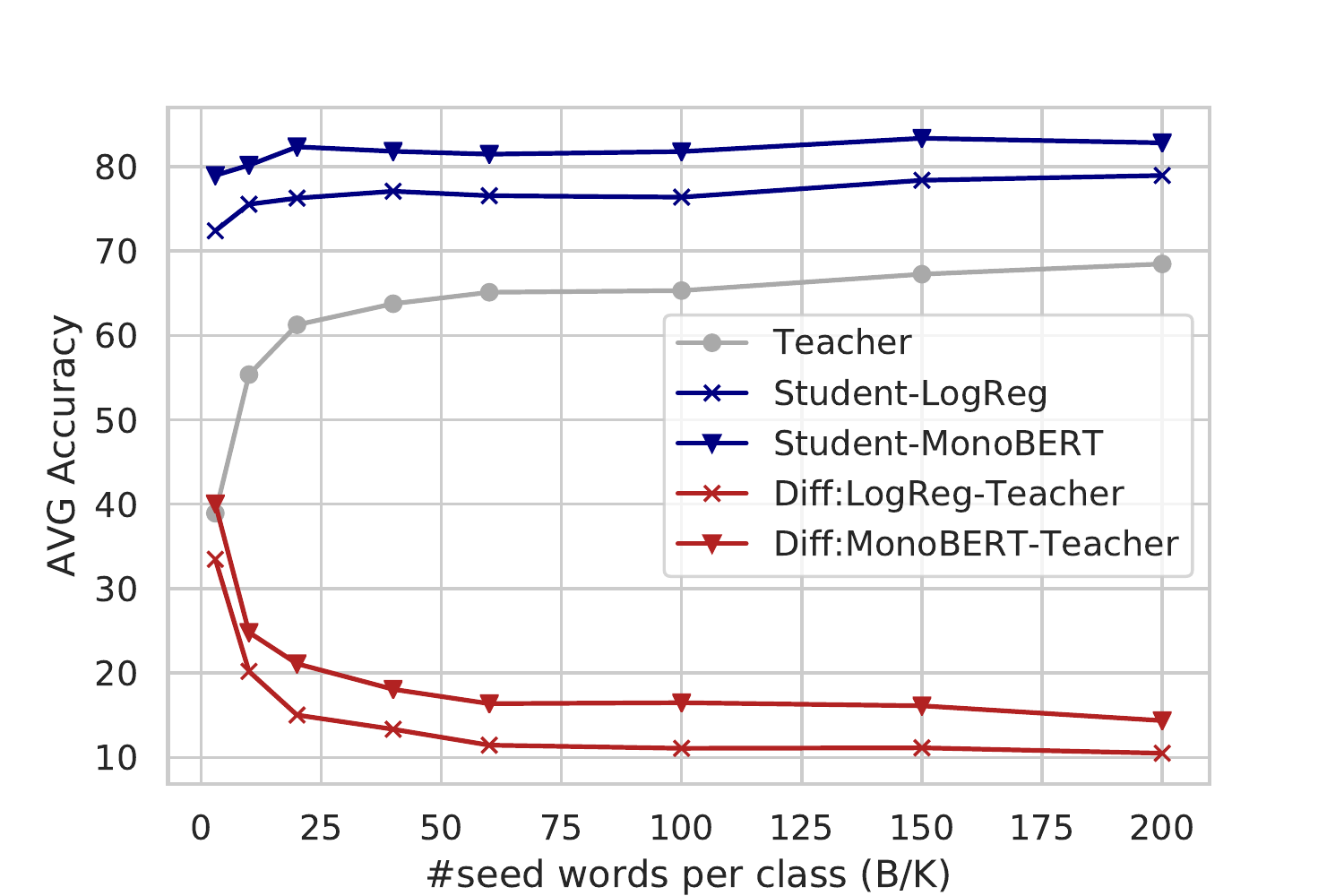}
	\caption{Validation accuracy across all MLDoc languages as a function of the translation budget $\frac{B}{K}$.}
\label{fig:mldoc_performance_seedwords}
\end{figure}

\paragraph{CLTS is effective under a minimal translation budget.}
Figure~\ref{fig:mldoc_performance_seedwords} shows CLTS's performance as a function of the number of seed words per class ($\frac{B}{K}$).
Even with just 3 seed words per class, Student-MonoBERT performs remarkably well.
Student's and Teacher's performance significantly increases with $\frac{B}{K}$ and most performance gains are obtained for lower values of $\frac{B}{K}$.
This is explained by the fact that CLTS prioritizes the most indicative seed words for translation. 
Therefore, as $\frac{B}{K}$ increases, the additional seed
words that are translated are less indicative than the already-translated seed words and as a result have lower chances of translating to important seed words in the target language.
The gap between the Teacher and Student performance has a maximum value of 40 absolute accuracy points and decreases as Teacher considers more seed words but does not get lower than 10, highlighting that Student learns predictive patterns in $L_T$ that may never be considered by Teacher. %

\begin{figure}
    \centering
    \includegraphics[scale=0.5]{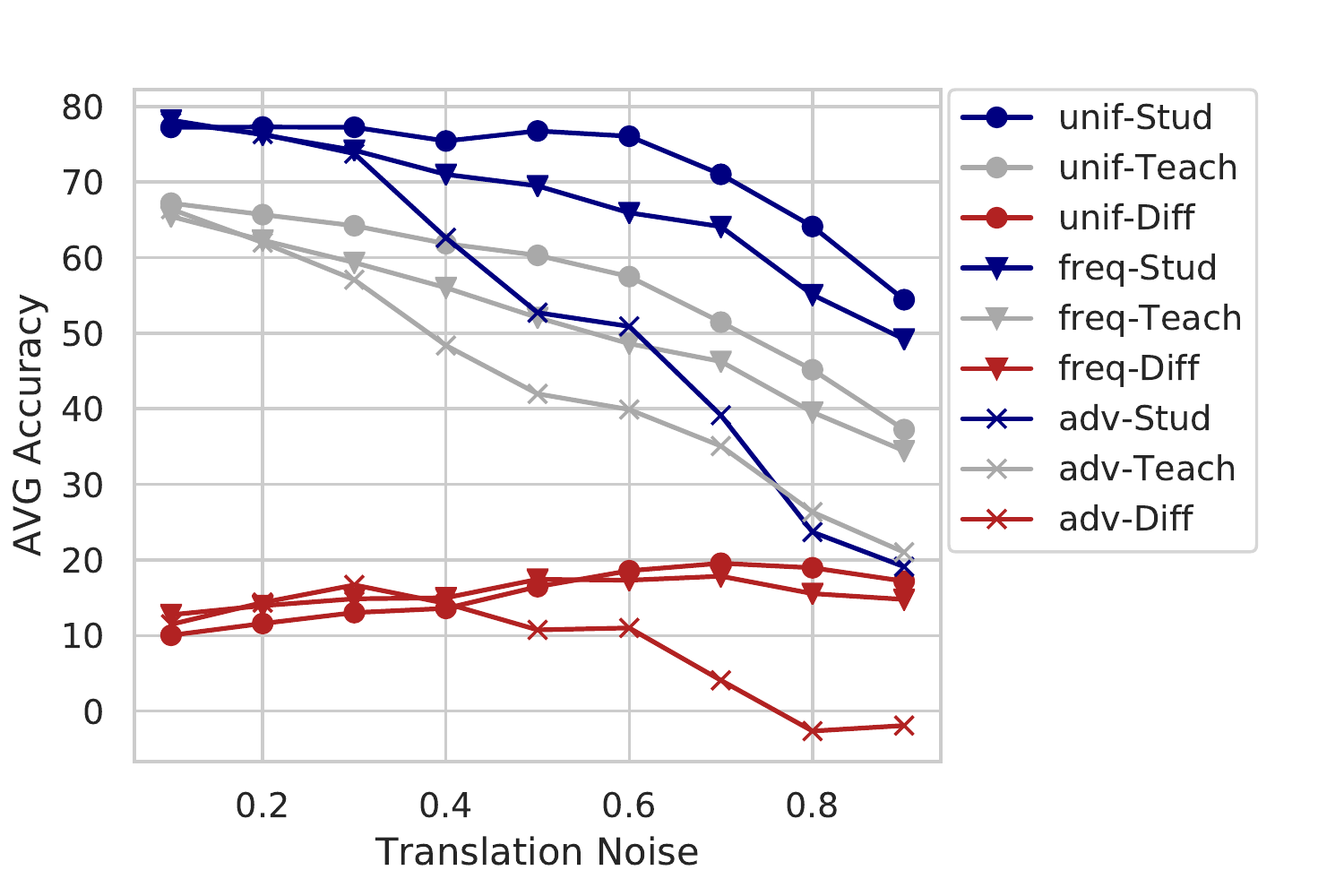}
    \caption{Average validation accuracy in MLDoc for Teacher (Teach), Student-LogReg (Stud), and their absolute difference in accuracy (Diff) under different scales of noise applied to the translated seed words: ``unif'' replaces a seed word with a different word sampled uniformly at random from $V_T$, ``freq'' replaces a seed word with a word randomly sampled from $V_T$ with probability proportional to its frequency in $D_T$, ``adv'' assigns a seed word to a different random class $k'\neq k$ by swapping its class weights in $\hat Z$.}
    \label{fig:mldoc_robustness_translation_noise_types}
\end{figure}

\paragraph{CLTS is robust to noisy translated seed words.}
In practice, an indicative seed word in $L_S$ may not translate to an indicative word in $L_T$. 
Our results above show that Student in CLTS performs well even when seed words are automatically translated across languages.
To further understand our method's behavior with noisy translated seed words, we introduce additional simulated noise of different types and severities.
According to Figure~\ref{fig:mldoc_robustness_translation_noise_types}, ``unif'' and ``freq'' noise, which replace translated seed words with random words, affect CLTS less than ``adv'' noise, which introduces many erroneous teacher-labels. 
Student is less sensitive than Teacher to noisy seed words: their performance gap (*-Diff) increases with the magnitude of translation noise (up to 0.7) for both ``unif'' and ``freq'' noise. 
Student's accuracy is relatively high for noise rates up to 0.3, even with ``adv'' noise: CLTS is effective even when 30\% of the translated seed words are assumed indicative for the wrong class.

\begin{figure}
    \centering
    \includegraphics[width=\columnwidth]{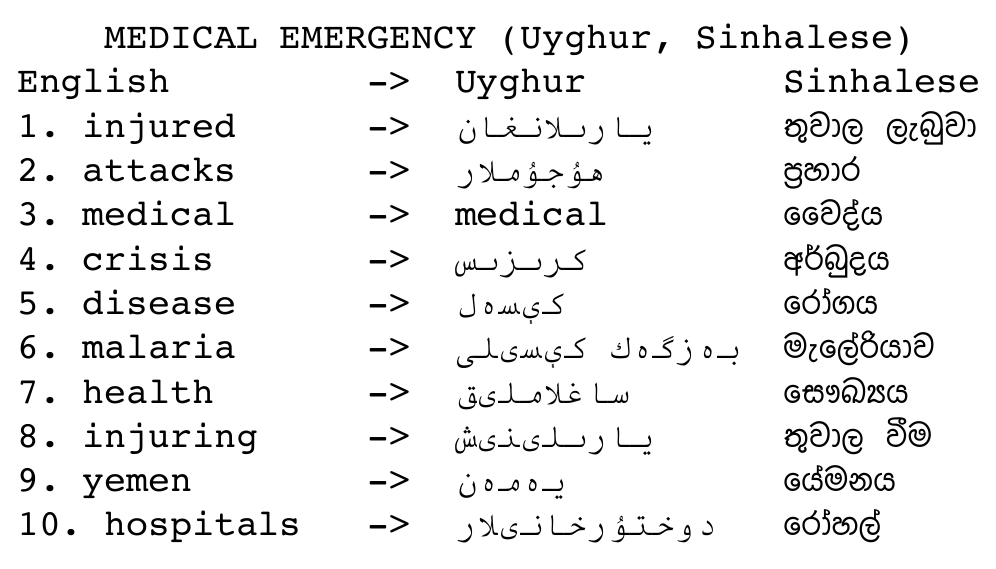}
    \caption{Top 10 extracted seed words for the ``medical emergency'' class and their translations to Uyghur and Sinhalese. Google Translate erroneously returns ``medical'' as a Uyghur translation of the word ``medical.''}
    \label{fig:lorelei-seedwords-small}
\end{figure}

\subsection{Addressing Emerging Classification Tasks in Low-Resource Languages}
We now show a preliminary exploration of CLTS for detecting medical emergency situations in the low-resource Uyghur and Sinhalese languages by just translating $B$=50 English seed words across languages. 
Figure~\ref{fig:lorelei-seedwords-small} shows the top 10 seed words transferred by CLTS for the medical emergency class.
We train Student-LogReg because BERT is not available for Uyghur or Sinhalese.
End-to-end training and evaluation of CLTS takes just 160 seconds for Uyghur and 174 seconds for Sinhalese.
The accuracy in Uyghur is 23.9\% for the teacher and 66.8\% for the student.
The accuracy in Sinhalese is 30.4\% for the teacher and 73.2\% for the student. 
The appendix has more details.
These preliminary results indicate that CLTS could be easily applied for emerging tasks in low-resource languages, for example by asking a bilingual speaker to translate a small number of seed words.
We expect such correct translations to lead to further improvements over automatic translations.  

\section{Conclusions and Future Work}
\label{s:conclusions}
We presented a cross-lingual text classification method, CLTS, that efficiently transfers weak supervision across languages using minimal cross-lingual resources.
CLTS extracts and transfers just a small number of task-specific seed words, and creates a teacher that provides weak supervision for training a more powerful student in the target language. 
We present extensive experiments on 4 classification tasks and 18 diverse languages, including low-resource languages.
Our results show that even a simple student outperforms the teacher and previous state-of-the-art approaches with more complex models and more expensive resources, highlighting the promise of generating weak supervision in the target language.
In future work, we plan to extend CLTS for handling cross-domain distribution shift~\cite{ziser2018deep} and multiple source languages~\cite{chen2019multi}. 
It would also be interesting to combine CLTS with available cross-lingual models, and extend CLTS for more tasks, such as cross-lingual named entity recognition~\cite{xie2018neural}, by considering teacher architectures beyond bag-of-seed-words. 

\subsubsection*{Acknowledgments}
We thank the anonymous reviewers for their constructive feedback. This material is based upon work supported by the National Science Foundation under Grant No. IIS-15-63785.

\newpage
\bibliography{anthology,emnlp2020}
\bibliographystyle{acl_natbib}

\newpage 
\newpage 
\appendix

\section{Appendix}
For reproducibility, we provide details of our implementation (Section~\ref{appendix-implementation-details}), datasets (Section~\ref{appendix-dataset-details}), and experimental results (Section~\ref{appendix-experiment-details}).
We will also open-source our Python code to help researchers replicate our experiments.

\subsection{Implementation Details}
\label{appendix-implementation-details}
We now describe implementation details for each component in CLTS: seed word extraction in $L_S$, seed word transfer, and teacher-student co-training in $L_T$.  

\paragraph{Source Seed Word Extraction:}
The inputs to the classifier in $L_S$ are tf-idf weighted unigram vectors\footnote{\url{https://scikit-learn.org/stable/modules/generated/sklearn.feature\_extraction.text.TfidfVectorizer.html}}. 
For the classifier, we use scikit-learn's logistic regression\footnote{\url{https://scikit-learn.org/stable/modules/generated/sklearn.linear\_model.LogisticRegression.html}} with the following parameters: penalty=``l1'', C=$\lambda_B$, solver=``liblinear'', multi\_class=``ovr''. 
In other words, we address multi-class classification by training $K$ binary ``one-vs.-rest'' logistic regression classifiers to minimize the $L1$-regularized logistic loss (LASSO).
(We use scikit-learn version 0.22.1, which does not support a ``multinomial'' loss with L1-penalized classifiers.) 
We tune $\lambda_B$ by computing the ``regularization path'' between $0.1$ and $10^7$, evenly spaced on a log scale into 50 steps.
To efficiently\footnote{Using a 16-core CPU machine, we compute $\lambda_B$ and train the source classifier in less than one minute (see Section~\ref{appendix-experiment-details}).} compute the regularization path, we use the ``warm-start'' technique~\cite{koh2007interior}, where the solution of the previous optimization step is used to initialize the solution for the next one. 
This is supported in scikit-learn by setting the warm\_start parameter of logistic regression to True. 

\paragraph{Seed Word Transfer:}
We obtain seed-word translations using the MUSE\footnote{\url{https://github.com/facebookresearch/MUSE\#ground-truth-bilingual-dictionaries}} bilingual dictionaries~\cite{conneau2017word}, which contain up to 100,000 dictionary entries per language pair. 
 Importantly, we use only the translations for $B\leq 500$ English seed words. 
 To understand the impact of translation budget in performance, we experiment with the following values for $\frac{B}{K}$: [2, 5, 10, 20, 30, 40, 50, 60, 70, 80, 90, 100, 150, 200]. We leave for future work the non-uniform distribution of seed words across classes, which might improve efficiency as ``easier'' classes may be modeled with fewer seed words.
 For Uyghur and Sinhalese, which have no entries in MUSE, we use Google Translate.
 For reproducibility, we cached the translations obtained from Google Translate and will share them with the code of the paper. 
If a source word has multiple translations in MUSE,\footnote{Various translations for a word in MUSE may correspond to different senses of the word. For example, the seed word ``shares'' for the ``Corporate'' topic translates to both ``comparte'' (share) and ``acciones'' (stocks) in Spanish.} we use all translations as noisy target seed words with the same weight,
while if a seed word has no translation in the target language, then we directly use it as a target seed word (this may be useful for named entities, emojis, etc.).
Translations provided by a human annotator would possibly lead to better target seed words but, as we show here, even noisy automatic translations can be effectively used in CLTS. 

\paragraph{Teacher-Student Co-Training:}
For the logistic regression (LogReg) student in $L_T$, we use scikit-learn's logistic regression with default parameters (including penalty=``l2'',  C=1). 
The inputs to LogReg are tf-idf weighted n-gram ($n$=1,2) vectors. 
For our monolingual BERT (MonoBERT) student, we use the following pre-trained models from huggingface\footnote{\url{https://huggingface.co}}:
\begin{itemize}
    \item English: \href{https://huggingface.co/bert-base-cased}{bert-base-cased}
    \item Spanish: \href{https://huggingface.co/dccuchile/bert-base-spanish-wwm-cased}{dccuchile/bert-base-spanish-wwm-cased}
    \item French: \href{https://huggingface.co/camembert-base}{camembert-base}
    \item German: \href{https://huggingface.co/bert-base-german-cased}{bert-base-german-cased}
    \item Italian: \href{https://huggingface.co/dbmdz/bert-base-italian-xxl-cased}{dbmdz/bert-base-italian-xxl-cased}
    \item Russian: \href{https://huggingface.co/DeepPavlov/rubert-base-cased}{DeepPavlov/rubert-base-cased}
    \item Chinese: \href{https://huggingface.co/bert-base-chinese}{bert-base-chinese}
    \item Japanese: \href{https://huggingface.co/cl-tohoku/bert-base-japanese}{bert-base-japanese}
\end{itemize}
We use the default hyperparameters in the ``Transformers'' library~\cite{Wolf2019HuggingFacesTS} and do not re-train (with the language modeling objective) MonoBERT in the target domain.
To avoid label distribution shift because of iterative co-training, we balance teacher-labeled documents in $D_T'$ by keeping the same number of documents across classes before training the student. 
We perform two rounds of teacher-student co-training, which has been shown to gain most of the improvement in~\citet{karamanolakis2019cotraining}. 
Table~\ref{tab:model parameters} reports the model parameters for each dataset and language. 
We do not tune any model hyperparameters and use default values instead.

\subsection{Dataset Details}
\label{appendix-dataset-details}
\paragraph{Document Classification in MLDoc:}
The Multilingual Document Classification Corpus (MLDoc\footnote{\url{https://github.com/facebookresearch/MLDoc}};~\citet{schwenk2018corpus}) contains Reuters news documents in English, German, Spanish, French, Italian, Russian, Chinese, and Japanese. 
Each document is labeled with one of the four categories: 
\begin{itemize}
    \item CCAT (Corporate/Industrial)
    \item ECAT (Economics)
    \item GCAT (Government/Social)
    \item MCAT (Markets)
\end{itemize}
MLDoc was pre-processed and split by~\citet{schwenk2018corpus} into 1,000 training, 1,000 validation, and 4,000 test documents for each language (Table~\ref{tab:mldoc-statistics}). 
We use labeled training documents only in English for training the source classifier. 
We treat training documents in German, Spanish, French, Italian, Russian, Chinese, and Japanese as unlabeled in CLTS by ignoring the labels. 

\paragraph{Review Sentiment Classification in CLS:} The Cross-Lingual Sentiment corpus (CLS\footnote{\url{https://webis.de/data/webis-cls-10.html}};~\citet{prettenhofer2010cross}) contains Amazon product reviews in English, German, French, and Japanese.
Each language includes product reviews from three domains: books, dvd, and music. 
Each labeled document includes a binary (positive, negative) sentiment label. Table~\ref{tab:cls-statistics} reports dataset statistics. 
Validation sets are not available for CLS.
We use labeled training documents only in English for training the source classifier. 
We ignore training documents in German, French, and Japanese, and use unlabeled documents in CLTS. 

\paragraph{Sentiment Classification in TwitterSent, Sentipers, and LORELEI:} The Twitter Sentiment corpus (TwitterSent;~\citet{mozetic2016}) contains Twitter posts in Bulgarian (Bg), German (De), English (En), Spanish (Es), Croatian (Hr), Hungarian (Hu), Polish (Pl), Portuguese (Pt), Slovak (Sk), Slovenian (Sl), and Swedish (Sv). 
We use the pre-processed and tokenized data provided by~\cite{rasooli2018cross}.
In addition to these tweets,~\citet{rasooli2018cross} also use pre-processed and tokenized Persian (Fa) product reviews from the SentiPers corpus~\cite{hosseini2018sentipers} and manually labeled Uyghur (Ug) documents from the LDC LORELEI corpus.
On the above datasets, each document is labeled with a sentiment label: positive, neutral, or negative. 
Table~\ref{tab:twittersent-statistics} reports dataset statistics. 
We use labeled training documents only in English for training the source classifier. 
We treat training documents in the rest of the languages as unlabeled. 

\begin{table}[]
    \centering
    \begin{tabular}{|l|c|c|c|}
    \hline
        \textbf{Language} &  \textbf{Train} & \textbf{Dev} & \textbf{Test} \\\hline 
        English (En) &  1,000 & 1,000 & 4,000 \\
        German (De) &  1,000 & 1,000 & 4,000 \\
        Spanish (Es) &  1,000 & 1,000 & 4,000 \\
        French(Fr) &  1,000 & 1,000 & 4,000 \\
        Italian (It) &  1,000 & 1,000 & 4,000 \\
        Russian (Ru) &  1,000 & 1,000 & 4,000 \\
        Chinese (Zh) &  1,000 & 1,000 & 4,000 \\
        Japanese (Ja) &  1,000 & 1,000 & 4,000 \\
        \hline 
    \end{tabular}
    \caption{MLDoc corpus statistics.}
\label{tab:mldoc-statistics}
\end{table}

\begin{table}[]
    \centering
    \resizebox{\columnwidth}{!}{
    \begin{tabular}{|l|c|c|c|c|}
    \hline
        \textbf{Language} & \textbf{Domain} & \textbf{Train} & \textbf{Unlabeled} & \textbf{Test} \\\hline 
        \multirow{3}{*}{English} & books &  2,000 & 10,000 & 2,000 \\
         & dvd &  2,000 & 10,000 & 2,000 \\
         & music &  2,000 & 10,000 & 2,000 \\\hline 
        \multirow{3}{*}{German} & books &  2,000 & 30,000 & 2,000 \\
         & dvd &  2,000 & 30,000 & 2,000 \\
         & music &  2,000 & 30,000 & 2,000 \\\hline 
        \multirow{3}{*}{French} & books &  2,000 & 30,000 & 2,000 \\
         & dvd &  2,000 & 16,000 & 2,000 \\
         & music &  2,000 & 30,000 & 2,000 \\\hline 
        \multirow{3}{*}{Japanese} & books &  2,000 & 30,000 & 2,000 \\
         & dvd &  2,000 & 9,000 & 2,000 \\
         & music &  2,000 & 30,000 & 2,000 \\ 
        \hline 
    \end{tabular}}
    \caption{CLS corpus statistics.}
\label{tab:cls-statistics}
\end{table}

\begin{table}[]
    \centering
    \begin{tabular}{|l|c|c|c|}
    \hline
        \textbf{Language} &  \textbf{Train} & \textbf{Dev} & \textbf{Test} \\\hline 

Arabic & - & 671 & 6100\\
Bulgarian & 23985 & 2999 & 2958\\
German & 63748 & 7970 & 7961\\
English & 46645 & 5832 & 5828\\
Spanish & 137205 & 17152 & 17133\\
Persian & 15000 & 1000 & 3027\\
Croatian & 56368 & 7047 & 7025\\
Hungarian & 36224 & 4528 & 4520\\
Polish & 116241 & 14531 & 14517\\
Portuguese & 63082 & 7886 & 7872\\
Russian & 44780 & 5598 & 5594\\
Slovak & 40476 & 5060 & 5058\\
Slovenian & 74268 & 9285 & 9277\\
Swedish & 32601 & 4076 & 4074\\
Uyghur & - & 136 & 346\\
\hline 
    \end{tabular}
    \caption{Twittersent, SentiPers, and LORELEI corpus statistics.}
\label{tab:twittersent-statistics}
\end{table}

\paragraph{Medical Emergency Situation Classification in LORELEI:}
The Low Resource Languages for Emergent Incidents (LORELEI) corpus~\cite{strassel2016lorelei} contains (among others) documents in Uyghur (Ug)\footnote{LDC2016E57\_LORELEI\_Uyghur} and Sinhalese (Si)\footnote{LDC2018E57\_LORELEI\_Sinhalese}.
Each document is labeled with an emergency need. Similar to~\citet{yuan2019interactive}, we consider binary classification to medical versus non-medical emergency need.  
Unfortunately, our number of labeled documents for each language is different than that reported in~\citet{yuan2019interactive}. 
In English, we use 806 labeled documents for training the source classifier. 
In Uyghur, we use 5,000 unlabeled documents for training the student and 226 labeled documents for evaluation. 
In Sinhalese, we use 5,000 unlabeled documents for training the student and 36 labeled documents for evaluation. 
Given the limited number of labeled documents, we do not consider validation sets for our experiments. 

\subsection{Experimental Result Details}
\label{appendix-experiment-details}
We now discuss detailed results on each dataset. 
In addition to baselines reported in the main paper, we also report supervised classifiers (*-sup) that were trained on each language separately using the labeled training data, to get an estimate for the maximum achievable performance.
We run CLTS 5 times using the following random seeds: [7, 20, 42, 127, 1993] and report the average performance results and the standard deviation across different runs.  
(The standard deviation for our LogReg student is negligible across all datasets so we do not report it.)
We report the results for the configuration of $B$ that achieves the best validation performance (accuracy for MLDoc, macro-average F1 for TwitterSent) and also report the validation performance, when a validation set is available. 

Table~\ref{tab:mldoc-results-full-all} reports results on MLDoc. 
\citet{eisenschlos2019multifit} report two different results for LASER~\cite{artetxe2019massively}: LASER-paper are the results reported in~\cite{artetxe2019massively}, while LASER-code are different results using the most recent LASER code. 
Here, we report both. (In Table~\ref{tab:mldoc-results-full}, we have reported the LASER configuration that achieves the best performance for each language.)
As expected, the performance of supervised models that consider in-language training datasets is higher than cross-lingual models. 

Table~\ref{tab:cls-results-full-all} reports results on CLS per domain. 
(In Table~\ref{tab:cls-results-full}, we reported the average performance across domains for each language.)
Note that MultiFiT-sup has substantially higher accuracy than MonoBERT-sup and LogReg-sup. 
This indicates that MulfiFit is probably a better model for this task. 
It would be interesting to evaluate in the future whether using MultiFiT as student outperforms Student-MonoBERT.

Table~\ref{tab:twittersent-results-all} reports results on TwitterSent, SentiPers, and LORELEI. We have reported the best performing approaches in~\citet{rasooli2018cross} that use En as a source language.
We noticed that CLTS achieves best validation performance using more seed words in the Twitter corpora compared to the MLDoc and CLS corpora.
We hypothesize that because Twitter posts are shorter than news documents or reviews, the context of seed words is less rich in indicative words and so the student requires larger teacher-labeled datasets to be effective.
Note, however, that even with a tighter budget of $B$=60, CLTS-Student has an average accuracy of 40.5\% and outperforms previous approaches relying on dictionaries or comparable corpora.

\begin{figure*}
    \centering
    \includegraphics[scale=0.9]{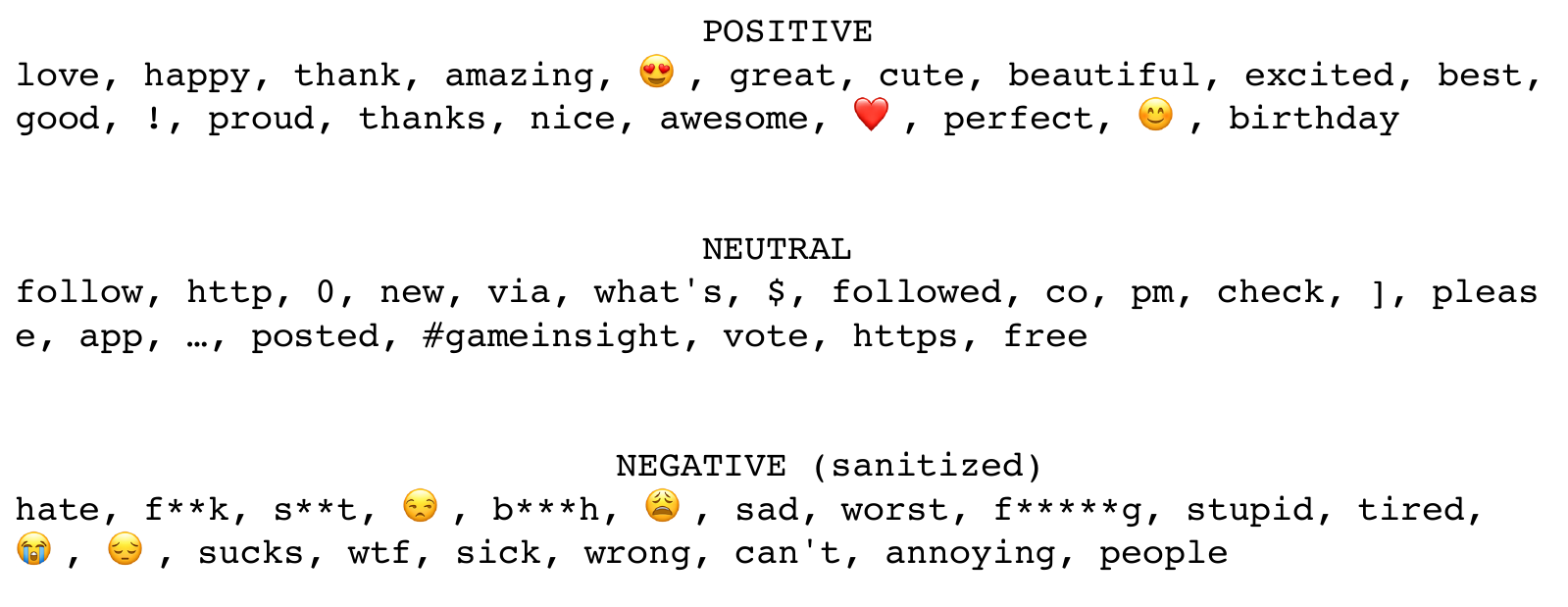}
    \caption{TwitterSent: Top 20 seed words extracted per class (Section~\ref{s:source-extraction}). Interestingly, some of the seed words are actually not words but emojis used by Twitter users to indicate the corresponding sentiment class.}
    \label{fig:twitter-seedwords}
\end{figure*}

\begin{figure*}
    \centering
    \includegraphics[scale=0.9]{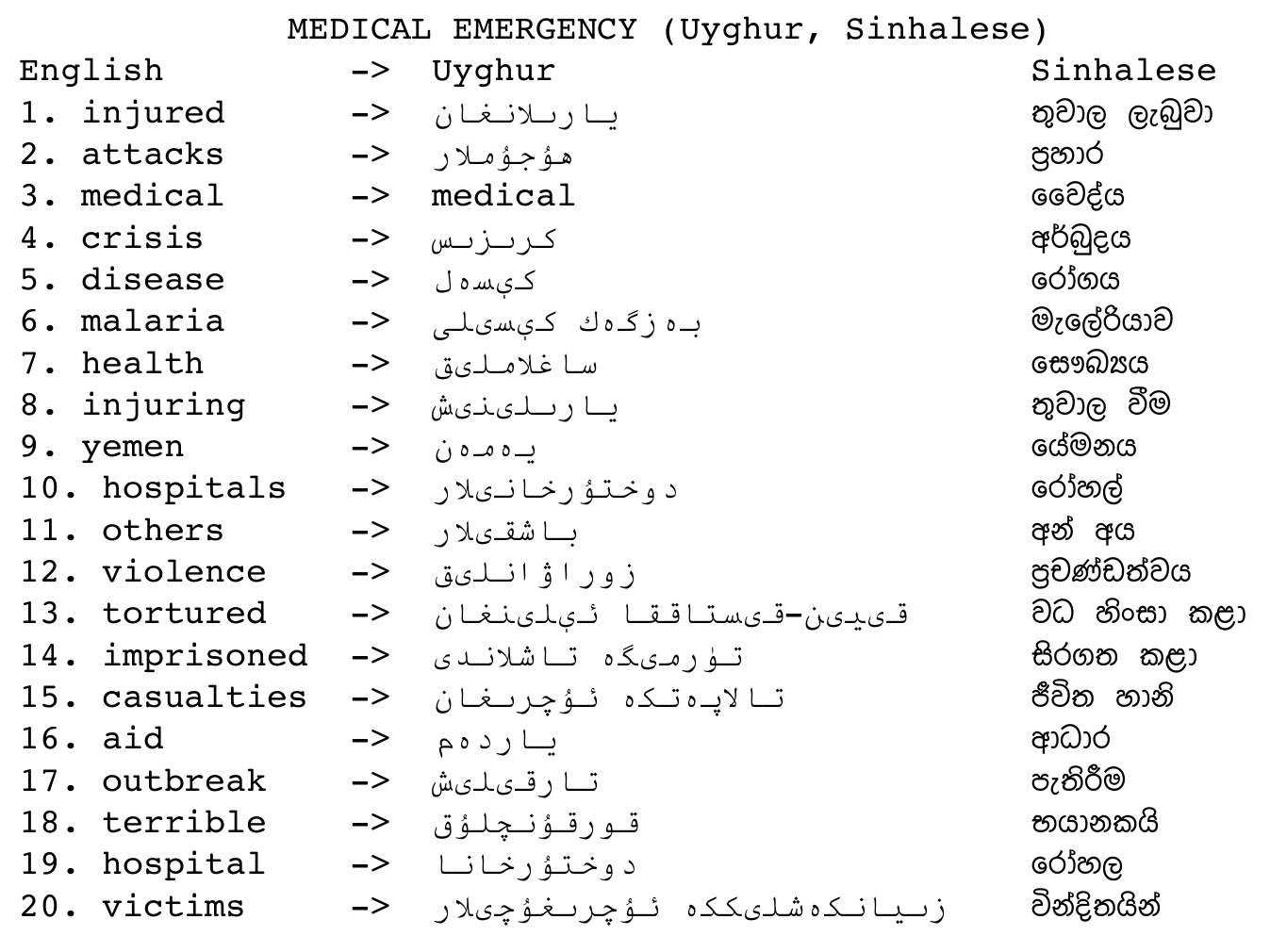}
    \caption{LORELEI: Top 20 seed words for the ``medical emergency'' class and their translations obtained through Google Translate. The incorrect translation for the important ``medical'' seed word from English to Uyghur is ``medical.'' }
    \label{fig:lorelei-seedwords}
\end{figure*}

\paragraph{Examples of Extracted Seed Words:}
Table~\ref{tab:mldoc-seedwords} reports the 10 most important seed words extracted for each of the four news document classes in CLS. 
Table~\ref{tab:cls-seedwords} reports the 10 most important seed words extracted for each binary class and domain in CLS.  
Figure~\ref{fig:twitter-seedwords} reports the 20 most important seed words extracted for each of the 3 sentiment classes in TwitterSent, SentiPers and LORELEI. 
Figure~\ref{fig:lorelei-seedwords} reports the 20 most important seed words extracted for the medical situation class in LORELEI and their translations to Uyghur and Sinhalese. 

\paragraph{Testing CLTS in Non-English Source Languages:}
To evaluate whether our results generalize to non-English source languages, we run additional experiments using De, Es, and Fr as source languages in CLS. 
For those experiments, we also consider En as a target language. 
Table~\ref{tab:nonenglish-source-langs} reports the evaluation results. 
Across all configurations, there is no clear winner between MultiCCA and MultiBERT, but our Student-LogReg consistently outperforms both approaches, indicating that CLTS is also effective with non-English source languages. 

\paragraph{Ablation Study:}
Table~\ref{tab:ablation-experiments} reports results on MLDoc by changing parts of CLTS. 
The first row reports Student-Logreg without any changes.
\textbf{Change (a):} using the clarity-scoring (similar to tf-idf weighting) method  of~\cite{angelidis2018summarizing} leads to 3\% lower accuracy than extracting seed words from the weights of a classifier trained through sparsity regularization. 
\textbf{Change (b):} obtaining translations through Google Translate leads to 0.8\% lower accuracy than using bilingual MUSE dictionary.
We observed that Google Translate sometimes translates words to wrong translations without extra context, while MUSE dictionaries provide more accurate translations. 
\textbf{Change (c):} updating Teacher similar to~\citet{karamanolakis2019cotraining}, where the Teacher updates seed word qualities but does not consider documents without seed words during training, leads to 1.3\% lower accuracy than our approach, which replaces the teacher by the student and thus considers even documents without seed words. 
\textbf{Change (d):} removing seed words from Student's input leads to 2.8\% lower accuracy than letting Student consider both seed words and non-seed words.
This shows that even without using seed words, Student still performs accurately (77.2\% accuracy across languages), indicating that Student successfully exploits indicative features in the context of the seed words.

\paragraph{Runtime:}
Table~\ref{tab:runtimes} reports the end-to-end runtime for each experiment (i.e., the total time needed to run the script), which includes: loading data, training, and evaluating CLTS. 
The runtime does not include dataset pre-processing, which was performed only once. 
We ran all experiments on a server with the following specifications: 16 CPUs, RAM: 188G, main disk: SSD 1T, storage disk: SDD 3T, GPU: Titan RTX 24G.

\begin{table*}[]
    \centering
    \begin{tabular}{|l|l|}
    \hline
       CCAT  &  company, inc, ltd, corp, group, profit, executive, newsroom, rating, shares\\\hline
        ECAT & bonds, economic, deficit, inflation, growth, tax, economy, percent, foreign, budget\\\hline
        GCAT & president, police, stories, party, sunday, people, opposition, beat, win, team\\\hline
        MCAT & traders, futures, dealers, market, bids, points, trading, day, copper, prices\\\hline
    \end{tabular}
    \caption{MLDoc: Top 10 English seed words extracted per class (Section~\ref{s:source-extraction}).}
    \label{tab:mldoc-seedwords}
\end{table*}

\begin{table*}[]
    \centering
    \resizebox{2\columnwidth}{!}{
    \begin{tabular}{|l|l|}
    \hline
       DVD-POS  &  best, great, excellent, love, highly, enjoy, wonderful, life, good, favorite\\\hline
        BOOK-POS & excellent, great, lives, wonderful, life, fascinating, fun, easy, love, best\\\hline
        MUSIC-POS & amazing, highly, great, favorites, best, favorite, awesome, classic, excellent, love\\\hline
       DVD-NEG & waste, boring, worst, bad, disappointing, disappointed, awful, poor, horrible, terrible\\\hline
    BOOKS-NEG & money, disappointed, disappointing, boring, disappointment, worst, waste, bad, finish, terrible\\\hline
    MUSIC-NEG & boring, worst, disappointment, poor, sorry, garbage, money, disappointing, bad, horrible\\\hline
    \end{tabular}}
    \caption{CLS: Top 10 English seed words extracted per class and domain (Section~\ref{s:source-extraction}).}
    \label{tab:cls-seedwords}
\end{table*}

\begin{table*}[]
\resizebox{2\columnwidth}{!}{
\begin{tabular}{|l|l|ccccccc|c|}
\hline
\textbf{Method} & \textbf{Cross-Lingual} & \multicolumn{7}{c|}{\textbf{Language}}& \textbf{AVG}\\
                                                        &    \textbf{Resource}           & \textbf{De}   & \textbf{Es}   & \textbf{Fr}   & \textbf{It}   & \textbf{Ru}   & \textbf{Zh}   & \textbf{Ja}   & \textbf{Acc}  \\\hline 
\multicolumn{9}{c}{\textit{Methods below use labeled target documents (supervised)}}\\\hline 
LogReg-sup                                                          & -    & 93.7 & 93.8 & 91.6 & 85.2 & 83.7 & 87.6 & 88.4 & 89.1 \\
MultiBERT-sup        & -    & 93.3 & 95.7 & 93.4 & 88.0   & 87.5 & 89.3 & 88.4 & 90.8 \\
MultiFiT-sup        & -    & \textbf{95.9} & \textbf{96.1} & \textbf{94.8} & \textbf{90.3}   & \textbf{87.7} & \textbf{92.6} & \textbf{90.0} & \textbf{92.5} \\\hline

\multicolumn{10}{c}{\textit{Methods below use parallel corpora}}\\\hline 
LASER, paper & parallel corpora & 86.3 & \textbf{79.3} & 78.3 & 70.2 & 67.3 & 71.0  & 61.0 & 71.2 \\
LASER, code & parallel corpora & 87.7 & 75.5 & 84.0 & 71.2 & 66.6 & 76.7 & 64.6 & 75.2\\
MultiFiT & LASER & \textbf{91.6} &	79.1 &	\textbf{89.4}&	\textbf{76.0} &	\textbf{67.8} & \textbf{82.5} &	\textbf{69.6} &	\textbf{79.4} \\\hline
\multicolumn{10}{c}{\textit{Methods below use pre-trained multi-lingual language models}}\\\hline 
MultiBERT & - & 79.8 & 72.1 & 73.5 & 63.7 & 73.7 & 76.0 & 72.8 & 73.1 \\
ST-MultiBERT & MultiBERT&\textbf{90.0} & \textbf{85.3} & \textbf{88.4} & \textbf{75.2}  & \textbf{79.3} & \textbf{87.0} & \textbf{76.8}  & \textbf{83.1} \\\hline 
\multicolumn{10}{c}{\textit{Methods below use bilingual dictionaries (Student requires Teacher)}}\\\hline 
MultiCCA &  $B=20K$ & 81.2 & 72.5 & 72.4 & 69.4 & 60.8 & 74.7 & 67.6 & 71.2 \\
Teacher & MUSE ($B=160$) & 72.7 & 73.5 & 77.6 & 62.5 & 46.9 & 53.3 & 31.9 & 59.8 \\
Student-LogReg & Teacher & 87.4 & 86.0 & 89.1 & 70.5 & 71.9 & 82.4 & 68.8 & 79.4 \\
Student-MonoBERT & Teacher & \textbf{90.4} & \textbf{86.3} & \textbf{91.2} & \textbf{74.7} & \textbf{75.6} & \textbf{84.0} & \textbf{72.6} & \textbf{82.1}\\\hline
\multicolumn{10}{c|}{\textit{Below we report the standard deviation of test accuracies across 5 runs}}\\\hline
Student-MonoBERT &  & 0.5& 0.5 & 0.4 & 0.7 & 0.8 & 0.4 & 0.6 &  \\\hline 
\multicolumn{10}{c|}{\textit{Below we report validation accuracies}}\\\hline
Teacher & MUSE ($B=160$) & 72.9 & 74.1 & 79.5 & 59.5 & 54.8 & 65.7 & 49.0 & \\
Student-LogReg & Teacher & 86.5 & 88.4 & 88.5 & 70.9 & 73.2 & 82.3 & 67.7 & \\
Student-MonoBERT & Teacher & 89.8 & 88.2 & 91.6 & 75.2 & 76.9 & 84.2 & 71.1 & \\
\hline 
\end{tabular}}
\caption{Accuracy results on MLDoc.}
\label{tab:mldoc-results-full-all}
\end{table*}

\begin{table*}[]
\resizebox{2\columnwidth}{!}{
\begin{tabular}{|l|l|ccc|ccc|ccc|c|}
\hline
\textbf{Method}        & \textbf{Cross-Lingual}                                                                                       & \multicolumn{3}{c|}{\textbf{De}} & \multicolumn{3}{c|}{\textbf{Fr}} & \multicolumn{3}{c|}{\textbf{Ja}} & \textbf{AVG}  \\
             &                                                                                              \textbf{Resource}       & Books  & DVD   & Music & Books  & DVD   & Music & Books  & DVD   & Music &    \textbf{Acc}  \\\hline 
\multicolumn{12}{c}{\textit{Methods below use labeled target documents (supervised)}}\\\hline   
LogReg-sup    & - & 84.5   & 82.8  & 84.1  & 84.7   & 86.0  & 88.0  & 80.9   & 83.0  & 83.0  & 84.1 \\
MultiBERT-sup    & - & 86.1   & 84.1  & 82.0  & 86.2   & 86.9  & 86.7  & 80.9   & 82.8  & 80.0  &  84.0 \\
MonoBERT-sup    & - & 82.4 & 80.0 & 81.7 & 88.4 & 86.2 & 86.3 & \textbf{86.3} & 85.7 & 86.2 & 84.8\\

MultiFiT-sup    & - & \textbf{93.2}   & \textbf{90.5}  & \textbf{93.0}  & \textbf{91.3}   & \textbf{89.6}  & \textbf{93.4}  & \textbf{86.3}   & \textbf{85.8}  & \textbf{86.6}  & \textbf{90.0} \\\hline

\multicolumn{12}{c}{\textit{Methods below use parallel corpora or MT systems}}\\\hline   
MT-BOW       & GoogleTransl.            & 79.7   & 77.9  & 77.2  & 80.8   & 78.8  & 75.8  & 70.2   & 71.3  & 72.0  & 76.0 \\
BiDRL        & Google Transl. & 84.4   & \textbf{84.1}  & \textbf{84.7}  & 84.4   & \textbf{83.6}  & 82.5  & 73.2   & 76.8  & 78.8  & 81.3 \\
CLDFA & parallel corpora & 84.0	& 83.1 &	79.0 &	83.4 &	82.6 & 	83.3&	77.4&	\textbf{80.5} &	76.5& 81.1\\
LASER, code        & parallel corpora                                                                & 84.2   & 78.0  & 79.2  & 83.9   & 83.4  & 80.8  & 75.0   & 75.6  & 76.3  & 79.5 \\
MultiFiT     & LASER  & \textbf{89.6}   & 81.8  & 84.4  & \textbf{87.8}   & 83.5  & \textbf{85.6}  & \textbf{80.5}   & 77.7  & \textbf{81.5}  & \textbf{83.6} \\\hline 
\multicolumn{12}{c}{\textit{Methods below use bilingual dictionaries or no cross-lingual systems}}\\\hline   
VECMAP & - & 76.0 &	76.3 &	73.5 &	77.8 &	78.6 &	78.1 &	55.9 &	57.6 &	54.4 & 69.8\\
MultiBERT    & - & 72.2   & 70.1  & 73.8  & 75.5   & 74.7  & 76.1  & 65.4   & 64.9  & 70.3  & 71.4 \\
CL-SCL       & $B=450$ pivots              & 79.5   & 76.9  & 77.8  & 78.5   & 78.8  & 77.9  & 73.1   & 71.1  & 75.1  & 76.5 \\
Teacher      & MUSE ($B=20$)                                                                                       & 42.1   & 36.0  & 36.3  & 47.9   & 51.6  & 46.2  & 17.9   & 23.9  & 26.2  & 36.5 \\
Student-LogReg   & Teacher & 76.0   & 77.8  & 82.2  & 78.8   & 80.0  & 80.1  & \textbf{77.2}   & \textbf{79.8}  & \textbf{78.9}  & 79.0 \\
Student-MonoBERT & Teacher                                                                                    & 77.9   & \textbf{79.9}  & \textbf{82.5}  & \textbf{84.3}   & 83.9  & \textbf{82.0}  & 76.4   & 77.7  & 78.8  & \textbf{80.4}\\\hline
\multicolumn{12}{c|}{\textit{Below we report the standard deviation of test accuracies across 5 runs}}\\\hline
Student-MonoBERT & Teacher &0.6   &  0.9& 0.8 & 0.9 &  0.4 & 0.5& 0.5& 0.4 & 0.2  &\\\hline
\end{tabular}}
\caption{Accuracy results on CLS. Validation accuracy is not reported as there is no validation set.}
\label{tab:cls-results-full-all}
\end{table*}

\begin{table*}[]
\resizebox{2\columnwidth}{!}{
\begin{tabular}{|l|l|cccccccccccccc|c|}
\hline 
Method & CL Resource & \textbf{Ar}   & \textbf{Bg}   & \textbf{De}   & \textbf{Es}   & \textbf{Fa}   & \textbf{Hr}   & \textbf{Hu}   & \textbf{Pl}   & \textbf{Pt}   & \textbf{Ru}   & \textbf{Sk}   & \textbf{Sl}   & \textbf{Sv}  & \textbf{Ug} & \textbf{AVG}  \\\hline
\multicolumn{17}{c}{\textit{Methods below labeled target documents (supervised)}}\\\hline 
LogReg-sup	& - & - &	54.4 &	54.4 &	43.8&	65.9&	56.0&	51.4&	56.4&	49.9&	56.6&	66.0&	57.0&	60.6 & - & - (56.0)\\
LSTM-sup&	-	& -&54.5&	59.9&	45.4&	67.8&	61.6&	60.4&	64.5&	51.1&	69.2&	70.1&	58.6&	62.5 & -  & - (60.5)\\\hline 
\multicolumn{17}{c}{\textit{Methods below use parallel corpora}}\\\hline 
CLWE-BQ                  & parallel corpora & 37.3 & 33.0   & 43.5 & 42.6 & 40.1 & 30.8 & 41.1 & 41.7 & 38.6 & 44.8 & 22.6 & 32.2 & 39.1 & 30.0 &37.0 \\
Project-LDC   & parallel corpora& 37.2 & -&-& 42.7 & 33.1 &-& 47.0   &- &-& 48.0   &-&      &- & 38.6 &- (41.1)\\
Project-EP & parallel corpora & - &\textbf{38.7} &	\textbf{47.3} &	41.8 &	- &	-& 38.1 &	38.8 &	\textbf{39.3} &-&\textbf{30.0}&	\textbf{44.6}&	\textbf{44.6}&-&(40.4)\\\hline 
\multicolumn{17}{c}{\textit{Methods below use comparable corpora or dictionaries}}\\\hline 
SentiWordNet               & SentiWordNet (>20K) & 25.6 & 30.6 & 32.0   & 25.3 & 25.3 & 19.8 & 29.2 & 26.0   & 22.9 & 29.5 & 19.2 & 28.1 & 22.7 & 36.7 & 26.6 \\
CLWE-Wikt    & Wiktionary (>20K) & 31.0   & 45.3 & 51.0   & 37.7 & 31.7 & -    & 40.8 & 32.9 & 35.4 & \textbf{43.8} & 36.6 & 32.1 & 40.4 &   28.0 &- (37.4) \\
CLWE-CP    & comparable corpora & 21.1 & 28.6 & 37.7 & 27.7 & 20.7 & 13.9 & 22.4 & 30.2 & 22.2 & 25.3 & 24.6 & 25.3 & 31.1 & 25.7 & 25.5 \\
Teacher       &  $B=500$& 22.7 & 42.8 & 45.5 & 42.7 & 30.9 & 36.4 & 39.4 & 40.7 & 34.4 & 29.8 & 40.4 & 29.5 & 38.7 & 20.3 &35.3 \\
Student-LogReg & Teacher & \textbf{39.0}   & \textbf{46.3} & \textbf{52.5} & \textbf{44.9} & \textbf{45.7} & \textbf{39.4} & \textbf{45.2} & \textbf{45.4} & 38.7 & 43.2 & \textbf{43.3} & \textbf{42.1} & \textbf{50.4} & \textbf{41.2} &\textbf{44.1}\\\hline 
\multicolumn{17}{c}{\textit{Below we report validation accuracies}}\\\hline 
Teacher &  $B=500$& 31.3 & 43.8 & 45.7 & 43.2 & 32.3 & 34.3 & 39.4 & 41.1 & 35.0 & 27.2 & 40.5 & 29.7 & 40.5& 22.8 & \\
Student-LogReg & Teacher &  47.2 & 48.7 & 52.1 & 45.4 & 46.5 & 39.0 & 46.9 & 45.3 & 40.1 & 41.7 & 43.2 & 42.5 & 50.0 & 38.3 &  \\

\hline
\end{tabular}}
\caption{Macro-averaged F1 results on TwitterSent, SentiPers, and LDC LORELEI.}
\label{tab:twittersent-results-all}
\end{table*}

\begin{table*}[]
\centering 
\resizebox{1.5\columnwidth}{!}{
\begin{tabular}{|c|cccc|}
\hline
 & \multicolumn{4}{c|}{\textbf{Target Acc} (MultiCCA / MultiBERT / Student-LogReg)}\\
Source Language  & En        & De        & Es        & Fr        \\\hline
En & - & 81.2/80.2/\textbf{87.4} & 72.5/76.9/\textbf{86.0}   & 72.4/72.6/\textbf{89.1} \\
De & 56.0/59.7/\textbf{82.8} & -         & 73.2/54.0/\textbf{81.3}   & 71.6/60.0/\textbf{84.9}\\
Es & 74.0/74.2/\textbf{80.8} &  55.8/57.6/\textbf{83.3} & - & 65.6/71.8/\textbf{89.0} \\
Fr & 64.8/76.1/\textbf{84.1} & 53.7/51.8/\textbf{84.5} & 65.4/72.1/\textbf{85.5} & -\\
\hline
\end{tabular}}
\caption{MultiCCA (left) vs. MultiBERT (center) vs. Student-LogReg (right) for various train (rows) and test (columns) configurations on MLDoc. Student-LogReg substantially outperforms MultiCCA and MultiBERT across all train and test configurations: CLTS effectively transfers weak supervision also from non-English source languages.}
\label{tab:nonenglish-source-langs}
\end{table*}

\begin{table*}[]
    \centering
    \resizebox{1.5\columnwidth}{!}{
    \begin{tabular}{|l|l|}
    \hline 
        \textbf{Change} & \textbf{AVG Acc} \\\hline%
        - (Original Student-LogReg) & \textbf{79.4} \\
    (a) Extract seed words as in~\citet{angelidis2018summarizing}  & 77.0 ($\downarrow$ 3.0\%)\\ %
     (b) Replace MUSE translations by Google Translate & 78.8 ($\downarrow$ 0.8\%)\\ %
     (c) Update Teacher as in~\citet{karamanolakis2019cotraining}  & 78.4 ($\downarrow$ 1.3\%)\\
     (d) Remove seed words from Student's input  & 77.2 ($\downarrow$ 2.8\%)\\\hline  %
    \end{tabular}}
    \caption{Ablation experiments on MLDoc.}
    \label{tab:ablation-experiments}
\end{table*}

\begin{table}[]
    \centering
    \resizebox{\columnwidth}{!}{
    \begin{tabular}{|c|c|c|c|}
    \hline 
         \textbf{Dataset}& \textbf{Lang} &  \textbf{LogReg} & \textbf{MonoBERT} \\\hline
   \multirow{7}{*}{MLdoc} & De & 14104 & 109M\\
        &Es & 15080 & 110M\\
        &Fr & 17632 & 111M\\
        &It & 11676 & 111M\\
        &Ru & 26804 & 178M\\
        &Zh & 15248 & 102M\\
        &Ja & 24676 & 111M\\\hline\hline
      \multirow{3}{*}{CLS-books}    & De &37560 &  109M \\
        & Fr &33462 &  111M\\
        & Ja &67195&  111M \\\hline
        \multirow{3}{*}{CLS-dvd}    & De &49832& 109M \\
        & Fr &12448 & 111M\\
        & Ja &61897 &  111M\\\hline
        \multirow{3}{*}{CLS-music}    & De &49899 & 109M\\
        & Fr &27194 & 111M\\
        & Ja &60554 & 111M\\\hline \hline

     \multirow{14}{*}{TwitterSent} &   Ar & 5502 & -\\
       & Bg& 44565 & -\\
       & De& 105993& -\\
        &Es& 245778& -\\
        &Fa& 44811& -\\
        &Hr& 108030& -\\
        &Hu& 50532& -\\
        &Pl& 184266& -\\
        &Pt& 83685& -\\
        &Ru& 58416& -\\
        &Sk& 76776& -\\
        &Sl& 140226& -\\
        &Sv& 70902& -\\
        &Ug& 978& -\\
        \hline\hline
    \multirow{2}{*}{LORELEI} &   Ug & 1353 & -\\
            &Si& 4654& -\\\hline 
    \end{tabular}}
    \caption{Number of model parameters for our LogReg and MonoBERT student in each dataset and language.}
    \label{tab:model parameters}
\end{table}

\begin{table}[]
    \centering
    \resizebox{\columnwidth}{!}{
    \begin{tabular}{|c|c|c|c|}
    \hline 
         \textbf{Dataset}& \textbf{Lang} &  \textbf{LogReg} & \textbf{MonoBERT} \\\hline
   \multirow{7}{*}{MLdoc} & De & 61s & 176s\\
        &Es & 33s & 165s\\
        &Fr & 43s & 139s\\
        &It & 29s & 157s\\
        &Ru & 54s & 195s\\
        &Zh & 70s & 173s\\
        &Ja & 51s & 170s\\\cline{2-4}
        &AVG & 49s & 168s\\\hline\hline
      \multirow{3}{*}{CLS-books}    & De &247s &  699s \\
        & Fr &301s &  837s\\
        & Ja &256s&  785s \\\hline
        \multirow{3}{*}{CLS-dvd}    & De &158s& 641s \\
        & Fr &71s & 277s\\
        & Ja &125s &  317s\\\hline
        \multirow{3}{*}{CLS-music}    & De &272s & 925s\\
        & Fr &290s & 884s\\
        & Ja &238s & 800s\\\cline{2-4}
        & AVG &218s & 685s\\
        \hline \hline
     \multirow{14}{*}{TwitterSent} &   Ar & 32s & -\\
       & Bg& 82s & -\\
       & De& 367s& -\\
        &Es& 2176s& -\\
        &Fa& 60s& -\\
        &Hr& 282s& -\\
        &Hu& 120s& -\\
        &Pl& 1445s& -\\
        &Pt& 361s& -\\
        &Ru& 164s& -\\
        &Sk& 181s& -\\
        &Sl& 654s& -\\
        &Sv& 145s& -\\
        &Ug& 20s& -\\\cline{2-4}
        &AVG& 434s& -\\
        \hline\hline
    \multirow{2}{*}{LORELEI} &   Ug & 160s & -\\
            &Si& 174s& -\\\cline{2-4}
            &AVG& 167s& -\\
            \hline 
    \end{tabular}}
    \caption{Runtimes for our LogReg and MonoBERT student in each dataset and language.}
    \label{tab:runtimes}
\end{table}

\end{document}